\documentclass[journal]{IEEEtran}

\usepackage{graphicx}
\usepackage{subfigure}
\usepackage{amsmath} % assumes amsmath package installed
\usepackage{amssymb}
\usepackage{accents}
\usepackage{multirow}
\usepackage{array}
\usepackage{tabularx}
\usepackage{booktabs}
\usepackage{threeparttable}
\usepackage[colorlinks,
            linkcolor=black,
            anchorcolor=black,
            citecolor=black]{hyperref}
\usepackage{color}

\begin{document}

\title{Contrastive Embedding Distribution Refinement and Entropy-Aware Attention for 3D Point Cloud Classification}

\author{Feng~Yang,
        Yichao~Cao,
        Qifan~Xue,
        Shuai~Jin,
        Xuanpeng~Li,
        and~Weigong~Zhang% <-this % stops a space
\thanks{This work was supported in part by the National Natural Science Foundation of China under Grant No. 61906038, the Fundamental Research Funds for the Central Universities under Grant No. 2242021R41184 , and the Zhishan Youth Scholar Program of Southeast University. We also thank the anonymous reviewers for their valuable comments and suggestions.}
\thanks{Feng Yang, Qifan Xue, Shuai Jin, Xuanpeng Li and Weigong Zhang are with School of Instrument Science and Engineering, Southeast University, Nanjing 210096, China (e-mail: yangfeng@seu.edu.cn).}
\thanks{Yichao Cao is with School of Automation, Southeast University, Nanjing 210096, China (e-mail: caoyichao@seu.edu.cn).}
\thanks{Corresponding Author: Xuanpeng Li, (e-mail: li\_xuanpeng@seu.edu.cn).}
}

\markboth{F.~Yang \lowercase{et al}.: Contrastive Embedding Distribution Refinement and Entropy-Aware Attention for 3D Point Cloud Classification}{F.~Yang et al.: Contrastive Embedding Distribution Refinement and Entropy-Aware Attention for 3D Point Cloud Classification}

\maketitle

\begin{abstract}
Learning a powerful representation from point clouds is a fundamental and challenging problem in the field of computer vision. Different from images where RGB pixels are stored in the regular grid, for point clouds, the underlying semantic and structural information of point clouds is the spatial layout of the points. Moreover, the properties of challenging in-context and background noise pose more challenges to point cloud analysis. One assumption is that the poor performance of the classification model can be attributed to the indistinguishable embedding feature that impedes the search for the optimal classifier. This work offers a new strategy for learning powerful representations via a contrastive learning approach that can be embedded into any point cloud classification network. First, we propose a supervised contrastive classification (SCC) method to implement embedding feature distribution refinement by improving the intra-class compactness and inter-class separability. Second, to solve the confusion problem caused by small inter-class variations between some similar-looking categories, we propose a confusion-prone class mining (CPCM) strategy to alleviate the confusion effect. Finally, considering that outliers of the sample clusters in the embedding space may cause performance degradation, we design an entropy-aware attention (EAA) module with information entropy theory to identify the outlier cases and the unstable samples by measuring the uncertainty of predicted probability. We assume that the outlier and unstable samples need to be ignored and enhanced, respectively, to enable the model to learn more robust decision boundaries.
The results of extensive experiments demonstrate that our method outperforms the state-of-the-art approaches by achieving 82.9\% accuracy on the real-world ScanObjectNN dataset and substantial performance gains up to 2.9\% in DCGNN, 3.1\% in PointNet++, and 2.4\% in GBNet. The code will be available at \textcolor{magenta}{\emph{https://github.com/YangFengSEU/CEDR}}.
\end{abstract}
%Formatter's Note: The journal limits the abstract to 200 words or less. Currently, your abstract contains 225 words. Please consider reducing the word count before submission to the journal.


\IEEEpeerreviewmaketitle

\begin{figure}
\centering
\includegraphics[width=9cm]{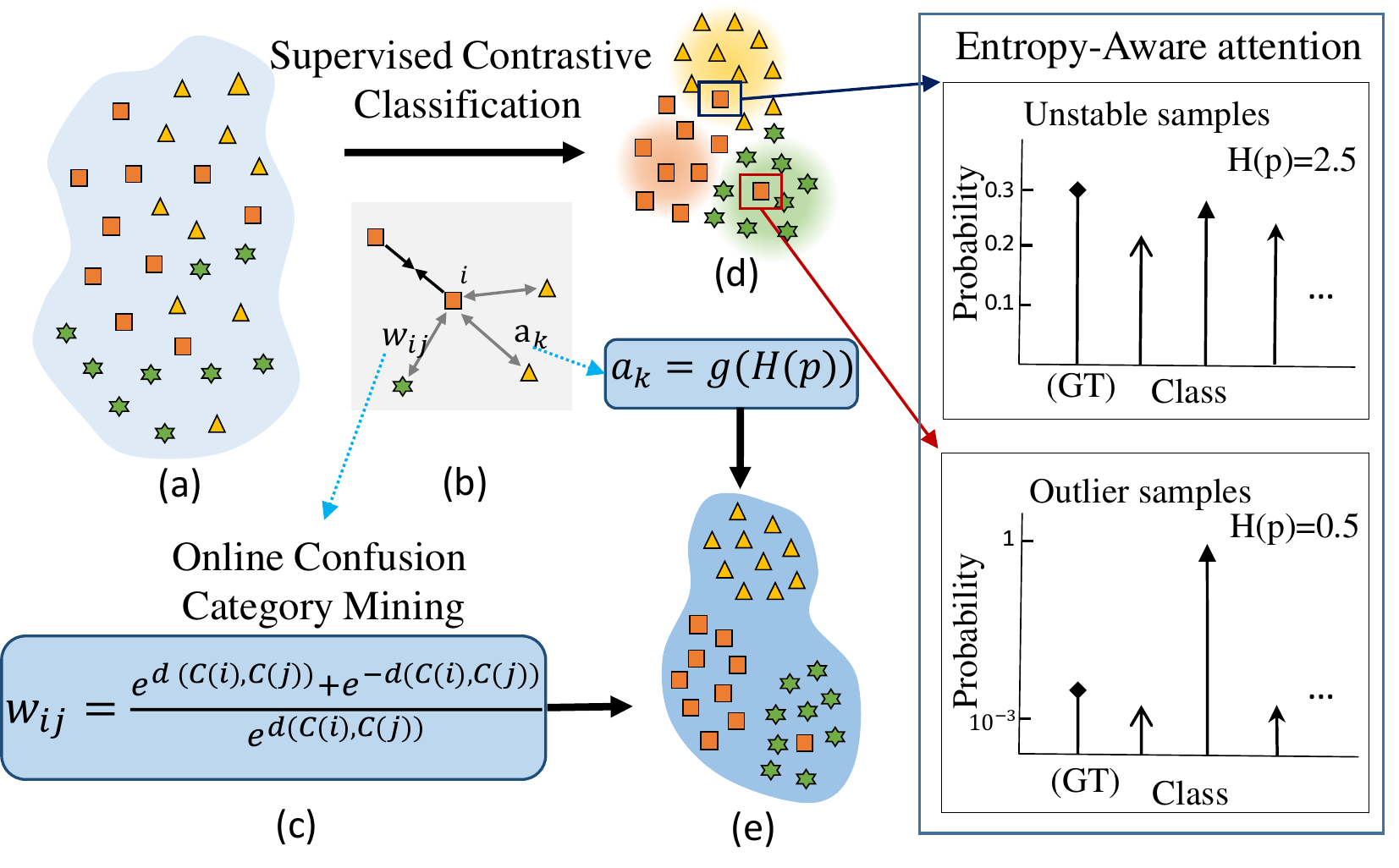}
\caption{Main idea. Current classification models learn to map input point clouds to an embedding space (a) while ignoring intrinsic semantic relationships of labeled data (\emph{i.e.}, inter-instance neighborhood relations of input space: whether semantically from a same class, noted with the same color). Supervised contrastive learning is introduced to foster a new training paradigm (b), by explicitly addressing intra-class compactness and inter-class separability. As shown in (b), each sample embedding feature $i$ is pulled closer ($\rightarrow$$\leftarrow$) to same class but pushed far (\textcolor[rgb]{0.498,0.498,0.498}{$\leftarrow$$\rightarrow$}) from other classes. We also apply the confusing-prone classes mining (CPCM) module to avoid confusion based on feature distance ($d(c(i),c(j)$) in (c)). In (d), we identify unstable samples (the uncertainty of the predicted probability distribution is high with the correct predict) and outlier samples (the uncertainty of predicted probability distribution is low with the incorrect predict) based on the information entropy. Thus a better-structured embedding space (e) is derived, eventually boosting the performance of the classification models.}
\end{figure}

\section{Introduction}

\IEEEPARstart{P}{oint} clouds are one of the most widely used 3D data representations in 3D scene understanding applications. Enabling machines to understand the 3D world is crucial for many important real-world applications, such as autonomous driving, augmented reality and robotics. Due to its tremendous contributions, point cloud analysis attracts much interest for research. A point cloud is a disordered set of points which can be generated from data that are captured with LiDAR or stereo/depth cameras. In contrast to images, in which RGB pixels are stored in a regular grid, for point clouds of 3D objects, the underlying semantic and structural information is the spatial layout of the points. Consequently, irregular point clouds cannot benefit like images from neighbor search by CNN's. In exception to the properties of irregularity and disorder, the challenging in-context and partial observations due to occlusions and reconstruction errors pose more challenges for real-world point cloud analysis (see Fig. 2 for detail).

Early pioneering works focused mainly on converting point clouds into other 3D data representations to extend deep learning methods from 2D image to 3D point cloud recognition tasks \cite{ref12}, such as multi-view projection \cite{ref13}\cite{ref14}\cite{ref15}\cite{ref16} and voxelization \cite{ref1}\cite{ref2}\cite{ref17}\cite{ref18}\cite{ref19}\cite{ref30}. PointNet \cite{ref20} provides a deep convolution framework that directly processes point sets with shared weights using multiple layer perceptron (MLP), which can handle unordered and unstructured 3D points. Many follow-up studies \cite{ref21}\cite{ref23}\cite{ref25}\cite{ref26}\cite{ref29} have made improvements. Most point-based state-of-the-art methods enhance model performance by incorporating low-level 3D relations (e.g., relative positions or distances) between the points into the network, which benefit from permutation invariance of low-level geometry \cite{ref24}\cite{ref27}\cite{ref28}.

However, due to the properties of challenging in-context and background noise, applying existing point cloud classification methods to real-world data may not produce the similar results as CAD models. As a result, high-dimensional features representing semantic information learned by the encoder may not be distinguishable enough that impede the search for the optimal classifier. (see Fig. 1(a)). Contrastive learning is a representation learning method that could refinement the distribution of the feature space to acquire a better semantic representation. It has been outstandingly successful in practice\cite{ref3}\cite{ref4}\cite{ref10}\cite{ref32}\cite{ref64}\cite{ref65}\cite{ref66} and involves learning transformation-invariant feature representations from unlabeled image data. The common strategy among those works is to pull an anchor towards a “positive” sample in the embedding space and to push the anchor away from many “negative” samples. Since no labels are available, a positive pair often consists of data augmentations that are formed by the anchors, and negative pairs are randomly chosen from the mini-batch.

\begin{figure*}
\centering
\includegraphics[width=13cm]{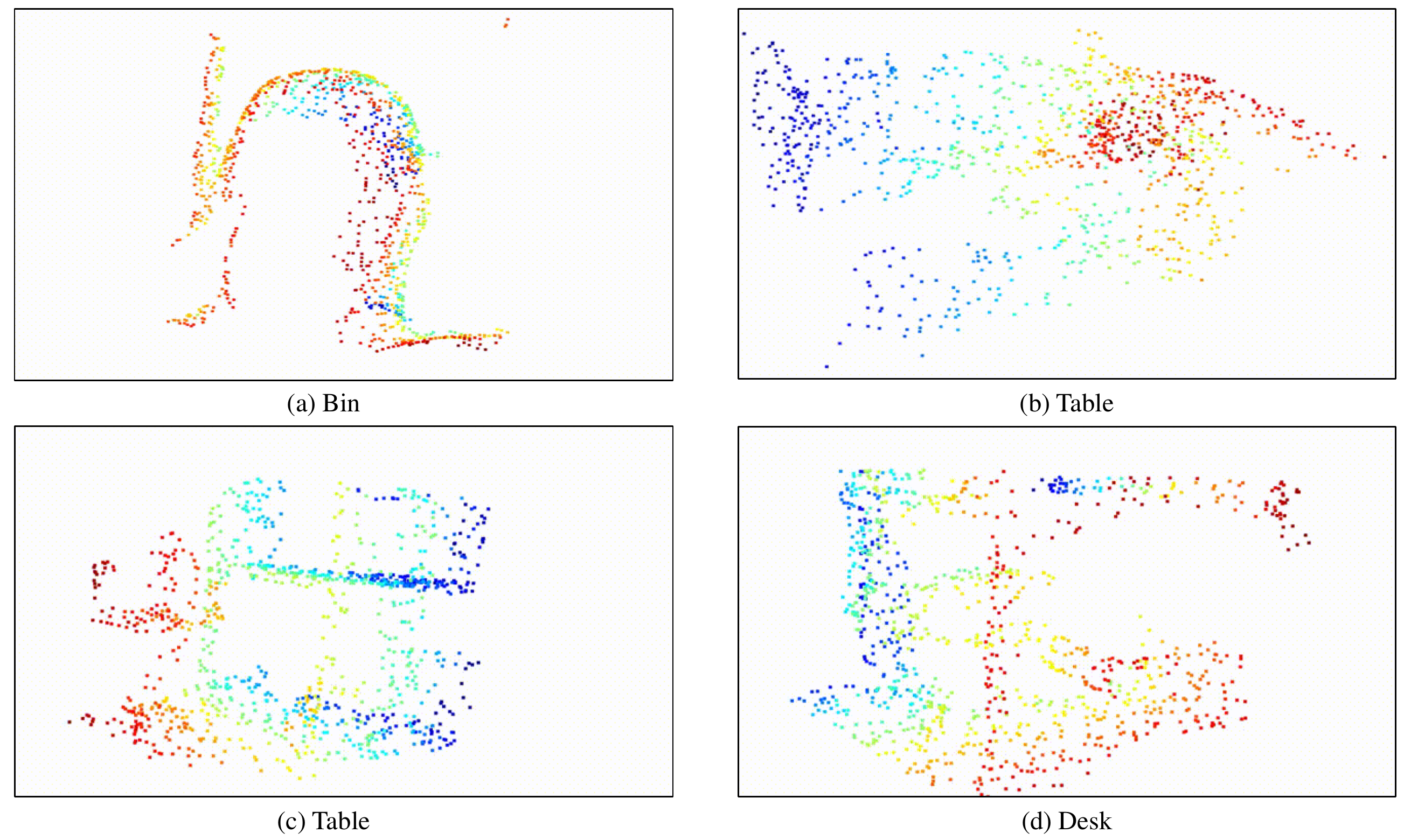}
\caption{The samples of real-world point clouds in ScanObjectNN. (a) is point clouds shaped of the bin which suffer from holes in the outer contour and background noise of wall and floor. (b) and (c) are both Tables but with large intra-class variance due to various designs. Conversely, The Desk (d) has a similar appearance (share chair feature) to (c), which causes a small inter-class variance between Desk and Table.}
\end{figure*}

For classification tasks, the cross-entropy loss is the most widely used in deep classification models. By assigning a target vector to each category, the deeply learned features are likely to be separable (Fig. 1(a))\cite{ref71}. However, for the point cloud recognition task, the deeply learned features must be not only separable but also discriminative (Fig. 1(e)). Point clouds that are extracted from real-world scenes may suffer from sparseness, lacking of contextual information, background noise, and inaccurate segmentation of point cloud objects. As a result, there is a large intra-class variance and a small inter-class variance in the real-world point cloud dataset. Hence, it is difficult to distinguish samples for softmax classifiers. Second, these methods with the softmax loss utilize only label information as a bootstrap signal. However, we believe that the data provide more valuable information than the labels. For example, an effective combination of self-supervised methods can exploit the neighborhood relations of the input space to learn discriminative features. Third, the training process of the model with softmax loss focuses only on the predictions of the ground-truth class. We believe that the probability distribution of the prediction vectors also contains valuable information for determining the reliability of the sample.

In this work, we propose a supervised contrastive classification for supervised learning that builds on the contrastive self-supervised literature by leveraging label information. As shown in Fig. 1(b), these positive samples are drawn from samples of the same class as the anchor to make use of the neighborhood relations of the input space rather than being data augmentations of the anchor, as in self-supervised learning. Consequently, normalized embeddings from the same class are pulled closer together than different classes, namely, we employ distribution refinement in the embedding space to promote the search for the optimal classifier. Ultimately, we designed a supervised contrastive loss and combined it with softmax loss via joint-supervision. To minimize the softmax loss, we try to map all samples within the same category to a single point (e.g. one-hot encoder) in feature space, which loses the intra-class variance. However, with the contrastive learning fashion, we can learn low-dimensional manifolds that preserve useful intra-class variances such as various designs. And we demonstrated an effective balance between supervised contrastive loss and softmax loss can boost the performance of the classification model. In our work, we consider many positive samples per anchor in addition to many negative samples, in contrast to self-supervised contrastive learning, which uses only a single positive sample. Our loss can be regarded as a generalization of both the triplet loss \cite{ref52} (only one positive sample and one negative sample per anchor) and N-pair losses \cite{ref74} (one positive sample and any number of negative samples).

Accordingly, we find that the combination of contrastive learning tends to confuse the classifier by pulling similar point cloud samples too close together (e.g., desk vs. table). In this paper, we refer to such samples as confusion-prone classes. To solve the problem of small inter-class variations between classes, we propose the confusing class pair mining (CPCM) module to avoid confusion by mining the distances between feature (Fig. 1(c)).

Furthermore, some samples are definitely incorrectly classified (low uncertainty of the predicted probability distribution). Some are indeterminately correctly predicted (high uncertainty of the predicted probability distribution). Such samples have a significant impact on the process of model optimization. The boundary assumption \cite{ref70} indicates that the decision boundary is more likely to be located in a low-density region of a cluster. Hence, nodes that lie in high-density regions are often reliable. Inspired by [70], we posit that the surely incorrectly predicted samples are likely to be located in high-density areas near the centers of clusters of other categories. Conversely, the indeterminately correctly predicted samples are likely to be located in low-density regions. As shown in Fig. 1(d), we refer to the surely incorrectly and indeterminately correctly predicted samples as outlier samples and unstable samples, respectively. Motivated by this observation, we assume that the unstable samples must be pulled apart to make the decision boundary clearer. At the same time, the outlier samples need to be ignored to avoid disturbing the gathering of samples of the same class. Our main contributions are summarized as follows:

\begin{itemize}
\item We propose a novel extension to the contrastive loss that can be embedded into any point cloud classification network. Our model can enhance the representative capabilities of deep models but does not need to modify the model architecture.
\item An online confusion-prone classes mining method is proposed to avoid the confusion among some classes. This phenomenon is usually caused by small inter-class variations between some similar classes. We alleviate this problem via metric learning on the embedding feature cluster distance.
\item We design entropy-aware attention to guide the model for dealing with unstable or outlier samples. These samples are critical to the process of model optimization.
\item Our method outperforms previous state-of-the-art methods by a large margin on a real-world 3D point cloud dataset, namely, ScanObjectNN. Our approach achieves a competitive 93.1\% accuracy, which represents an improvement of +2.4 compared to the original model, on the ModelNet40 dataset.
\end{itemize}

\section{Related Work}

As our method is most closely related to deep learning on 3D point clouds, contrastive learning and sample mining in deep metric learning, in what follows, we discuss representative works from each topic.

\subsection{Deep Learning on Point Clouds}
In contrast to image data, point clouds do not directly contain the spatial structure of a regular grid; they are collections of 3D point coordinates with an unordered layout. For deep models, feature learning for this data representation type must be considered while designing the network. Some methods attempt to convert the point cloud into an intermediate representation, such as a multiview image or volumetric data, to adapt convolution to the data.

A straightforward approach is to voxelize the point cloud in a 3D grid structure to form volumetric data \cite{ref1}\cite{ref2}\cite{ref7}. However, the representation is inefficient, as most voxels are usually unoccupied in 3D space. Later, OctNet \cite{ref30} was proposed, which explores the sparsity of voxel data to alleviate this problem. However, since voxels are discrete representations of space, this method still requires high-resolution grids with large memory consumption as a trade-off to maintain a suitable level of representation quality. Another common 3D representation is multiview representation \cite{ref12}\cite{ref13}\cite{ref14}\cite{ref15}\cite{ref16}], where the point data are projected to various image planes to form 2D images. By this approach, point data can benefit from conventional convolution on 2D images. However, this approach suffers from structural information loss during the projection, which causes feature quality reduction.

In \cite{ref20}, the irregular format and permutation invariance of point sets are discussed, and a network that consumes point clouds directly, namely, PointNet, is proposed. PointNet++ \cite{ref21} extends PointNet by further considering not only global information but also local details. The method uses the farthest sampling layer and a grouping layer to aggregate the local region information. Many follow-up studies have improved the aggregation of local region information \cite{ref22}\cite{ref23}\cite{ref25}\cite{ref26}\cite{ref29}\cite{ref31}. DGCNN \cite{ref29} aggregates the local context information with a dynamic graph network to link each center point with its k nearest neighbors. PointWeb \cite{ref22} gives attention to the interaction between points in each local neighborhood region and exhausts the context information between all point pairs. SRN \cite{ref23} found that the features of different point cloud patches can interact with each other due to the continuity and symmetry of the 3D object. Hence, the method designs a structural relationship module for aggregating local information based on point cloud patches. Subsequent works, such as GBNet \cite{ref31}, pointConv \cite{ref26}, and relation-shape CNN \cite{ref25}, also focus on the local structures of point clouds.

While recent works realize state-of-the-art point cloud deep learning by better aggregating the local region information, this work offers a new route for learning a powerful representation using contrastive learning. At a higher level, we directly employ feature distribution refinement to extract features, and our method is suitable for all the cited PointNet++ variants.

\subsection{Contrastive Learning}
Contrastive learning is a typical discriminative self-supervised learning \cite{ref3}\cite{ref4}\cite{ref9}\cite{ref32}\cite{ref34}\cite{ref34}\cite{ref43} method that aims to learn useful representations of the input data without relying on task-specific manual annotations. Recent advances in self-supervised visual representation learning based on contrastive methods show that self-supervised representations outperform their supervised counterparts \cite{ref5}\cite{ref34}\cite{ref35}\cite{ref36}\cite{ref37} on several downstream transfer learning benchmarks. This approach aims at embedding augmented versions of a sample close to each other while trying to push away embeddings from other instances. The difference between contrastive learning and generative self-supervised learning \cite{ref38}\cite{ref39}\cite{ref40} is that the former focuses on detailed features at the pixel level while the latter prefers to learn higher-level, semantically discriminative features.

Recent works argue that representations should additionally be invariant to unnecessary details and preserve as much information as possible. In \cite{ref50}, they refer to these two properties as alignment and uniformity and regard them as measures of representation quality. It also proved that the contrastive loss asymptotically optimizes these properties to prevent complete collapse whereby all representation vectors shrink into a single point. Various methods have been proposed for solving the collapsing problem, which rely on different mechanisms. Some approaches, such as instance discrimination \cite{ref44}, SimCLR \cite{ref6} and MoCo \cite{ref7}, define positive and negative sample pairs, which are treated differently in the loss function, and push the negative samples away to maintain the alignment properties. BYOL \cite{ref8} and SimSiam \cite{ref9} use stop gradients and an extra predictor to prevent collapse without negative pairs; SWAV \cite{ref10} uses an additional clustering step; and Barlow twins \cite{ref41} minimizes the redundant information between two branches.

Most works demonstrate that selecting robust pretext tasks along with suitable data augmentations can greatly boost the quality of representations. In the image domain, data augmentation involves mainly color transformation and geometric transformations such as cropping, resizing, rotation, and flipping \cite{ref3}\cite{ref6}\cite{ref45}\cite{ref46}\cite{ref47}\cite{ref48}. Recently, SwAV \cite{ref10} outperformed other self-supervised methods by using multiple augmentations. SimCSE \cite{ref6} has also achieved good results using dropout results for data augmentation in the NLP (natural language processing) field. Some recent works implement data augmentation by adversarial attacks to obtain positive and negative samples \cite{ref51}.

Contrastive learning in the image processing and NLP fields enables better use of data in a self-supervised fashion. Considering fully supervised classification tasks, the categories of samples naturally provide ``positive'' and ``negative'' samples instead of design pretext. For a better-structured embedding, ``positive'' samples (from the same class) needs to be pulled closer, while ``negative'' samples (from a different class) must be pushed away. Hence, we propose a novel extension of contrastive learning that enables better use of data via a self-supervised approach, thereby adapting contrastive learning to fully supervised point cloud classification. We utilize a SimClR-like approach that defines positive and negative sample pairs and treats them differently in the loss function such that the negative samples are pushed away to maintain the alignment properties.

\subsection{Sample Mining in Deep Metric Learning}
Closely related to contrastive learning is the family of losses that are based on the metric distance \cite{ref52}\cite{ref53}\cite{ref54}\cite{ref55}\cite{ref56}\cite{ref57}\cite{ref58}\cite{ref59}\cite{ref74}\cite{ref75}\cite{ref79}. Deep metric learning aims to learn a deep embedding that can capture the semantic similarity of data points by constructing pairs and introducing relative constraints among these pairs. Deep metric learning methods include doublet-based methods with the contrastive loss \cite{ref54}\cite{ref58}\cite{ref59}, triplet-based methods with the triplet loss \cite{ref52}\cite{ref53}\cite{ref75} and quadruplet-based \cite{ref57} methods with the double-header hinge loss. The key distinction between deep metric losses and the contrastive learning loss is the numbers of positive and negative pairs. Our loss can be regarded as a generalization of both the triplet loss  \cite{ref52}(only one positive and one negative sample per anchor) and N-pair losses \cite{ref74} (one positive and any number of negative samples). Although we can construct many image pairs for deep metric learning, a large fraction of trivial pairs will contribute zero to the loss once the model reaches a reasonable performance. Thus, it is intuitive to mine nontrivial pairs during training to achieve faster convergence and better performance.

Sample mining has been widely explored, and a range of mining methods have been proposed \cite{ref52}\cite{ref55}\cite{ref56}\cite{ref57}\cite{ref58}\cite{ref59}\cite{ref60}. Depending on the distances between the anchor and the positive and negative samples, sample mining can be divided into mining of difficult negative samples \cite{ref53}\cite{ref55}\cite{ref56}, mining of semidifficult negative samples, and mining of simple negative samples \cite{ref52}\cite{ref60}. Mining of positive samples is applied in [56][56][57][58]. Mining of semipositive samples is explored in \cite{ref59}. Local positive sample mining \cite{ref60} is proposed for learning an extended manifold rather than a contracted hypersphere. Most of these works mine a subset of pairs by using a binary score for each pair (keeping it or dropping it), and our work conducts soft mining using a continuous score (different weights are assigned to different pairs). Furthermore, in contrast to most mining strategies that are designed for samples, our mining solution considers hard class pairs.

In addition, samples with larger training losses are prioritized when difficult pair mining is applied. Therefore, outliers are generally mined and perturb the model training due to their large losses. Semi-difficult negative sample mining \cite{ref52}\cite{ref60} and multilevel mining \cite{ref58} are applied to remove outliers. Similarly, many works \cite{ref76}\cite{ref77} select reliable samples based on sample confidence. Instead, we propose EAA with information entropy theory for measuring confidence, identifying outliers and assigning smaller weights to them.

\begin{figure*}
\centering
\includegraphics[width=18cm]{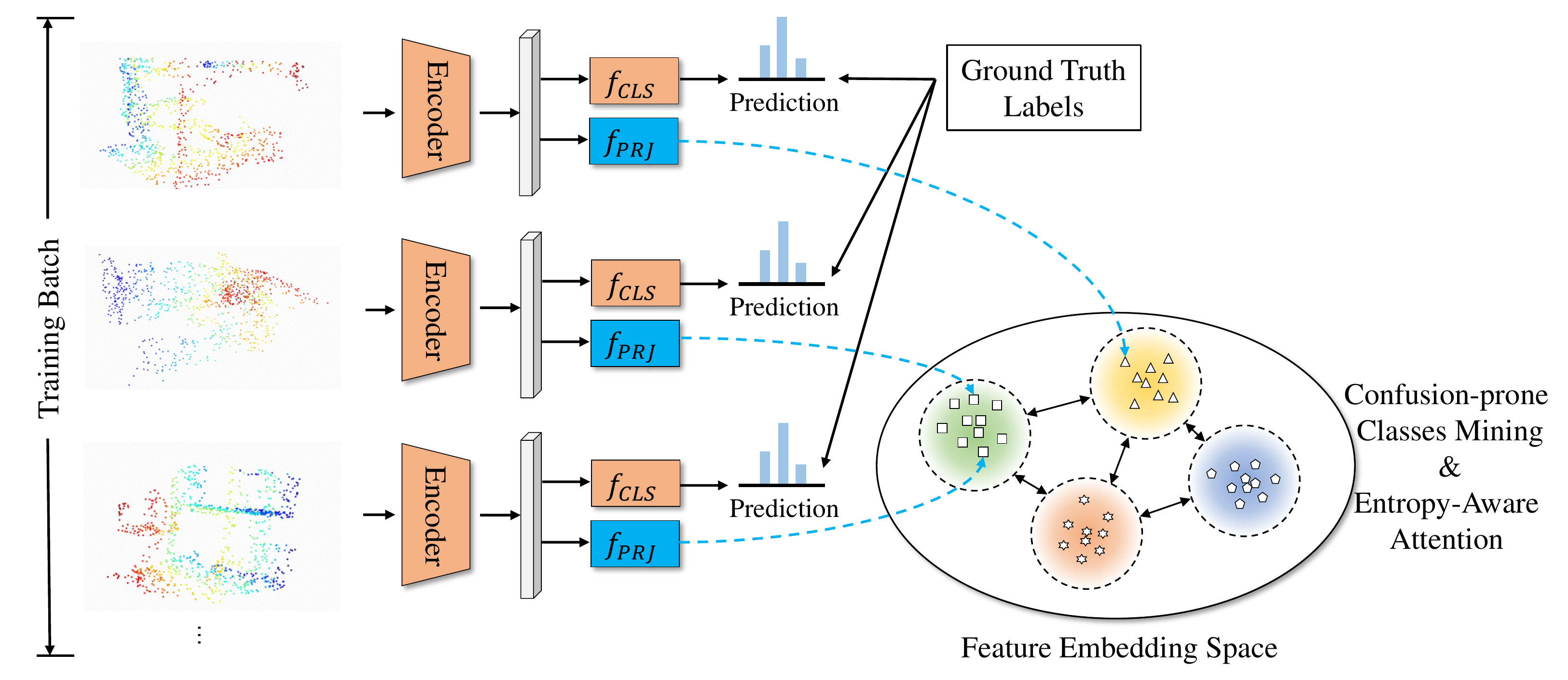}
\caption{Our network architecture. Generic point-based point cloud processing models are used as the encoder network. Firstly, one input point cloud data batch is transformed into a feature representation by the encoder network. Then, a classification head $f_{cls}$ and projection head $f_{prj}$ are designed for category prediction and feature embedding, respectively. In feature embedding space, we use a supervised contrastive classification module to get a better representation by pulling together the same class samples and pushing apart different class samples. Additionally, we introduce a confusing-prone classes mining strategy and an entropy-aware attention mechanism to further refine the high-level embedding distribution.}
\end{figure*}

\section{Method}
Fig. 3 presents the framework of the proposed method. As illustrated in Fig. 3, generic point-based point clouds analysis models are used as encoder networks. In the process of model optimization, one input point cloud data batch is transformed into a feature representation by the encoder network. Then, a classification head $f_{cls}$ and projection head $f_{prj}$ are designed for category prediction and feature embedding, respectively. Our method focuses on feature embedding distribution refinement and can be embedded into any point cloud classification network. We hope to learn more discriminative, generic and robust representations that can better reflect the underlying semantic and structural information of point cloud data.

In the following subsections, supervised contrastive learning for the point cloud class is introduced first. We describe a supervised contrastive paradigm for improving the intra-class compactness and inter-class separability of the feature embedding space. Second, we present the confusing-prone classes mining strategy, which enables the deep model to focus on easily confused categories. Finally, we describe the entropy-aware attention operation, which is used to better deal with unstable and outlier samples.

\subsection{Supervised Contrastive Classification}

Given a set of 3D points $P \subset \mathbb{R}^{3}$ with $N$ elements, $P_{i}$ represents the 3D coordinates of point $i$. The point cloud classification task aims to categorize a point cloud sample into a correct class $c \in \mathrm{C}$. Current approaches typically cast this classification task as an end-to-end scheme in which the cross-entropy loss is used in training. These methods generally obtain a global embedding $I \in \mathbb{R}^{D}$ with an aggregation encoder $f(\cdot)$(e.g., PointNet ++) and pass the embedding $I$ through a classification head $f_{cls}$ to predict the final categorical probability $y=f_{c l_{S}(I)}=\left|y_{1}, \cdots, y_{|C|}\right| \in[0,1]^{|C|}$. Given the predicted probability tensor $y$ and its ground-truth label $\bar{c} \in \mathrm{C}$, the cross-entropy loss is optimized:
\begin{equation}
\mathcal{L}_{p}^{C E}=-\log {y}_{\bar{c}}
\end{equation}

However, to minimize the cross-entropy loss, the model maps all samples with the same category to specified positions and dissimilar vectors to distant positions in the feature space without considering the diversity of the inter-class variance. Such a training paradigm mainly suffers from the isolation of individual samples: the loss function penalizes each prediction independently while ignoring the relationships between samples. More likely, the model tries to map all samples within the same category to a single point (\emph{e.g.} one-hot encoder) in feature space, which loses the intra-class variance. Consequently, the deep feature is separable but not discriminative enough to be used directly for recognition since the decision boundaries are hard to learn. In this work, we develop a supervised contrastive learning approach that makes full use of the global context of labeled point cloud datasets. In addition to using the one-to-one relationship between the samples and labels as a supervisory signal, we also explore the data distribution in input space, especially the cross-instance relationship among the categories.

Given a 3D point cloud $P$ with its ground semantic label $\bar{c}$, the positive samples $P^{+}$ are the samples belongs to the class $\bar{c}$, while the negative samples $P^{-}$ belonging to the other classes $C\verb|\|\bar{c}$. Assuming that there are $N$ samples from a batch, each is paired with others, thereby forming $N-1$ pairs for each anchor sample. We extend the popular contrastive loss function InfoNCE [3]to our supervised point cloud classification setting, which is defined as follows:
\begin{equation}
\small
\mathcal{L}_{\mathrm{p}}^{\mathrm{NCE}}=\frac{1}{\left|\mathcal{P}_{\mathrm{p}}\right|} \sum_{\mathrm{I}^{+} \in \mathcal{P}_{\mathrm{p}}}-\log \frac{\exp \left(\mathrm{I} \cdot \mathrm{I}^{+}\right)}{\exp \left(\mathrm{I} \cdot \mathrm{I}^{+}\right)+\sum_{\mathrm{I}^{-} \in \mathcal{N}_{\mathrm{p}}} \exp \left(\mathrm{I} \cdot \mathrm{I}^{-}\right)} {y}_{\overline{\mathrm{c}}}
\end{equation}
where $I$ denote a global embedding of anchor sample $P$, $P_{p}$ and $N_{p}$ represent paired collections of the positive and negative samples, respectively. The vector inner product represents the distance of the vector as a measure of similarity. To minimize the contrastive loss, the distances of the positive pairs in the embedding space are reduced, and those of the negative pairs are enlarged. Via this strategy, we learn an embedding space, where the same class samples are pulled together and different class samples are pushed apart.

Additionally, in this work, the contrastive loss and cross-entropy loss are complementary to each other; the latter forces the deep features of different classes to stay apart, which is meaningful for classification, while the former helps regularize the model with learning low-dimensional manifolds that preserve useful intra-class variances. Hence, we combine them by joint supervision, and the overall training target is as follows:
\begin{equation}
\mathcal{L}_{\text {overall }}=\sum_{p} \mathcal{L}_{p}^{C E}+\lambda \mathcal{L}_{\mathrm{p}}^{\mathrm{NCE}}
\end{equation}
Where $\lambda$ is a coefficient that is used to balance the two loss functions.

\subsection{Confusing-prone Classes Mining}
In the previous subsection, we developed an effective contrastive loss function to improve the intra-class compactness and inter-class reparability of the embedding space. However, this method still faces various problems. Some categories of samples are easily confused for similar appearance. For example, the desk and table categories in the ScanObjectNN dataset are difficult to distinguish for both the human eye and a deep model. The most likely reason is that they share similar geometric features, with table tops and table legs. From the perspective of the feature distribution in the embedding space, this phenomenon is also observed. As shown in Fig. 7, desk samples are represented as pink points, and the table samples are represented as red points. These two clusters are very close in the embedding space, and the classification boundary is not clear; as a result, they likely lead to substantial confusion for the classifier.

Motivated by this observation, we propose a novel confusing class mining module for enhancing the learning of confusing categories and further adjusting the distribution of the feature space to improve the classification performance, as illustrated in Fig. 4. To identify the categories that are easily confused, first, we calculate the mean feature of all samples that belong to the same category in the feature space as the center of the corresponding class. Then, we obtain the distance between each pair of class centers. We assume that the closer two clusters are, the more likely they are to be confused. Conversely, the farther apart the class centers are, the less attention the model needs to give to the distinction between these two types of features. Therefore, the similarity weight for classes $i$ and $j$ is defined as follows:
\begin{equation}
W_{i, j}^{-}=\frac{e^{\operatorname{dist}(C(i), C(j))}+e^{-\operatorname{dist}(C(i), C(j))}}{e^{\operatorname{dist}(C(i), C(j))}}
\end{equation}
Where $C(i)$ denote the feature center of class $i$, and $dist(C(i),C(j))$ is the feature center distance between class $i$ and $j$. Then, the rewighting for negative samples in CPCM is as follows:
\begin{equation}
L_{C P C M}(N)=\frac{\sum_{\left(x_{i}, x_{j}\right) \in N} W_{i, j}^{-} \exp \left(f\left(x_{i}\right)^{T} f\left(x_{j}\right)\right)}{\sum_{\left(x_{i}, x_{j}\right) \in N} W_{i, j}^{-}}
\end{equation}
Via this approach, we explicitly guide the model to focus on the distinction between confusion-prone categories.

\begin{figure}
\centering
\includegraphics[width=6cm]{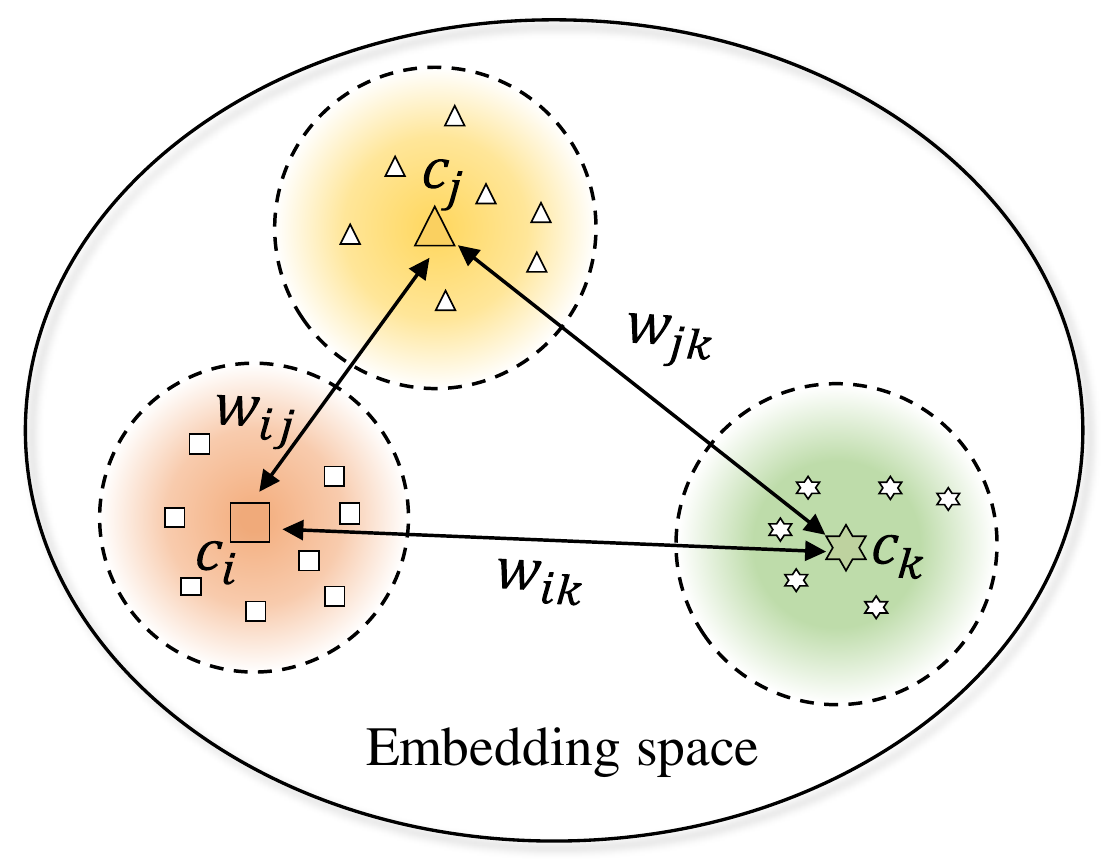}
\caption{Illustration of confusion-prone classes mining module (CPCM). The differently shaped dots indicate different categories of samples. The larger shaped dots represent the centers of the clusters. The forces that act on the points are represented by arrows (where each force acts on every point of the corresponding class, not only the center point as drawn). The length of each arrow represents the approximate strength of the force, where $W_{i j}>W_{j k}>W_{i k}$. The figure shows three categories, two of which are close together (yellow and pink) and one (green) that is farther away. In this module, the closer categories are pushed away more than the farther category.}
\end{figure}

\subsection{Entropy-Aware Attention}
In the previous subsection, we proposed a method for refining the feature distribution in high-dimensional feature space by the distances of clusters. However, we found that some inevitable unstable points and outliers in the model training have a bad impact on contrastive learning. Under these disturbances, increasing the complexity of the model may further aggravate overfitting. It is likely that the model unknowingly extracted some overdetailed features for the underlying generic representation.

In the process of model optimization, the features of some samples will deviate from the clusters that represent the correct semantic category and be closer to the other class clusters, which we call outliers in this paper. To minimize the contrastive loss, the samples with the same class are expected to be as compact as possible. Due to the existence of outliers, the clusters that are composed of the most correctly classified points will become sparse and deviate from the original cluster centers. In addition, the outliers will make the classifiers learn more complex decision boundaries and, thus, lead to overfitting.

To reduce the effects of outliers, we propose using information entropy for sample selection. Entropy is a measurable physical property that is most commonly associated with a state of disorder, randomness, or uncertainty. Ludwig Boltzmann explained entropy as the measure of the number of possible microscopic arrangements or states of individual atoms or molecules of a system that comply with the macroscopic condition of the system. The greater the number of possible microscopic arrangements or states, the higher the entropy.

In this work, the prediction probabilities of samples are regarded as the microscopic conditions, while the prediction accuracy is regarded as the macroscopic state. For a reliable node that is located in the low-density region of a cluster, the confidence in the class prediction that is made by the model should probably be high. The score of the cluster class is high, while the scores of the others are low, which leads to a small number of possible microscopic arrangements and a low entropy value. In contrast, for a less reliable node that is located near the decision boundary, the confidence in the class prediction should probably be low, and several categories are predicted with similar scores, thereby leading to a high entropy value. As shown in Fig. 6, we visualize the data distribution with various entropies. The Shannon entropy \cite{ref73} of a sample can be calculated as follows:
\begin{equation}
E_{P}=-\sum_{i=0}^{C} p\left(c_{i}\right) \log _{2} p\left(c_{i}\right); p\left(c_{i}\right)= \frac{c_{i}}{\sum_{0}^{C} c_{i}}
\end{equation}
Where $p(c_{i})$ denotes the normalized score of each category predicted by the deep model. $E_{P}$ denoted the entropy score of sample $P$.

\begin{figure}
\centering
\includegraphics[width=6cm]{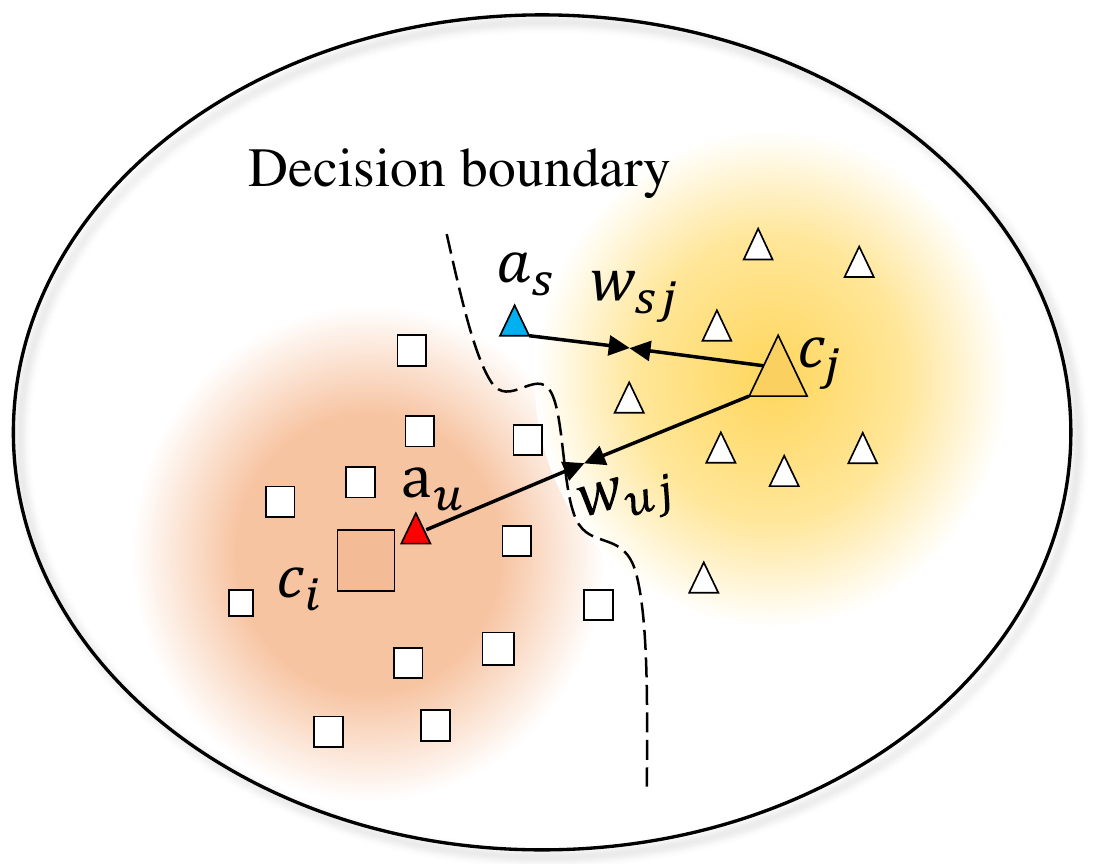}
\caption{Illustration of Entropy-aware attention module (EAA). The differently shaped dots indicate different categories. The larger shaped dots represent the centers of the clusters. The forces that act on the points are represented by arrows (where each force acts on every point of the corresponding class, not only the center point as drawn). The red node indicates an outlier near the center of the other class, and the blue node indicates an unstable point near the decision boundary, where $a_{s}$,$a_{u}$ are as introduced in Eq.7, and $w_{sj}$,$w_{uj}$ are as introduced in Eq.8.}
\end{figure}

Consequently,we can define outlier samples and unstable samples using the information entropy. Samples that have a low entropy value (this is an experience value, which we set to less than 1) and misclassified samples are regarded as outliers. At the same time, we correctly classify samples with a high entropy value (greater than 2.5) as unstable samples, which needs attention.

We propose a novel weighted contrastive loss for learning a more targeted discriminative embedding, as illustrated in Fig. 5. Specifically, we set different weights for outlier and unstable nodes in the contrastive learning loss, which are defined as follows:
\begin{equation}
a_{P}=\left\{\begin{array}{cc}
E_{P} ; & p \subset O u t l i e r s \\
E_{P}-1.2 ; & p \subset U n s t a b l e {N o d e s }
\end{array}\right.
\end{equation}
Where $E_{P}$ is entropy value of sample $P$. As a result, the weights of the outliers are less than 1, while those of the unstable nodes are greater than 1.2 (which is an experience value). To ensure that the unstable samples receive more attention than normal data nodes, the outlier samples are ignored whenever possible. When applied to the contrastive loss, the selection function of the sample pair weights is defined as follows:
\begin{equation}
\small
W_{i j}=\operatorname{select}\left(a_{i}, a_{j}\right)=\left\{\begin{array}{lr}
\max \left(a_{i}, a_{j}\right) ; & a_{i}, a_{j} \geq 1 \\
\min \left(a_{i}, a_{j}\right) ; & a_{i} \geq 1 \text { and } a_{j} \leq 1 \\
\min \left(a_{i}, a_{j}\right) ; & a_{i} \leq 1 \text { and } a_{j} \geq 1 \\
\min \left(a_{i}, a_{j}\right) ; & a_{i}, a_{j} \leq 1
\end{array}\right.
\end{equation}

In Eq. (8), we assign a small weight when one of the two samples is an outlier, and when neither is an outlier, the larger is used. The contrastive loss of the EAA module is as follows:
\begin{equation}
L_{E A A}(N)=\frac{\sum_{\left(x_{i}, x_{j}\right) \in N} W_{i, j}^{-} \exp \left(f\left(x_{i}\right)^{T} f\left(x_{j}\right)\right)}{\sum_{\left(x_{i}, x_{j}\right) \in N} W_{i, j}^{-}}
\end{equation}
\begin{equation}
L_{E A A}(P)=\frac{\sum_{\left(x_{i}, x_{j}\right) \in P} W_{i, j}^{+} \exp \left(f\left(x_{i}\right)^{T} f\left(x_{j}\right)\right)}{\sum_{\left(x_{i}, x_{j}\right) \in P} W_{i, j}^{+}}
\end{equation}

In our entropy-aware attention mechanism, the prediction probability entropy is used to identify outlier and unstable samples. We aim to separate the outlier data nodes and the most correctly classified points by penalizing the learning process of the outliers. In addition, we argue that unstable samples in sparse areas need to be pulled apart to make the classification boundary clearer. As a result, the optimal classifier search is greatly simplified, and the convergence is substantially accelerated. The relevant experimental results will be presented in the Experiment section.

\begin{figure}
\centering
\includegraphics[width=9cm]{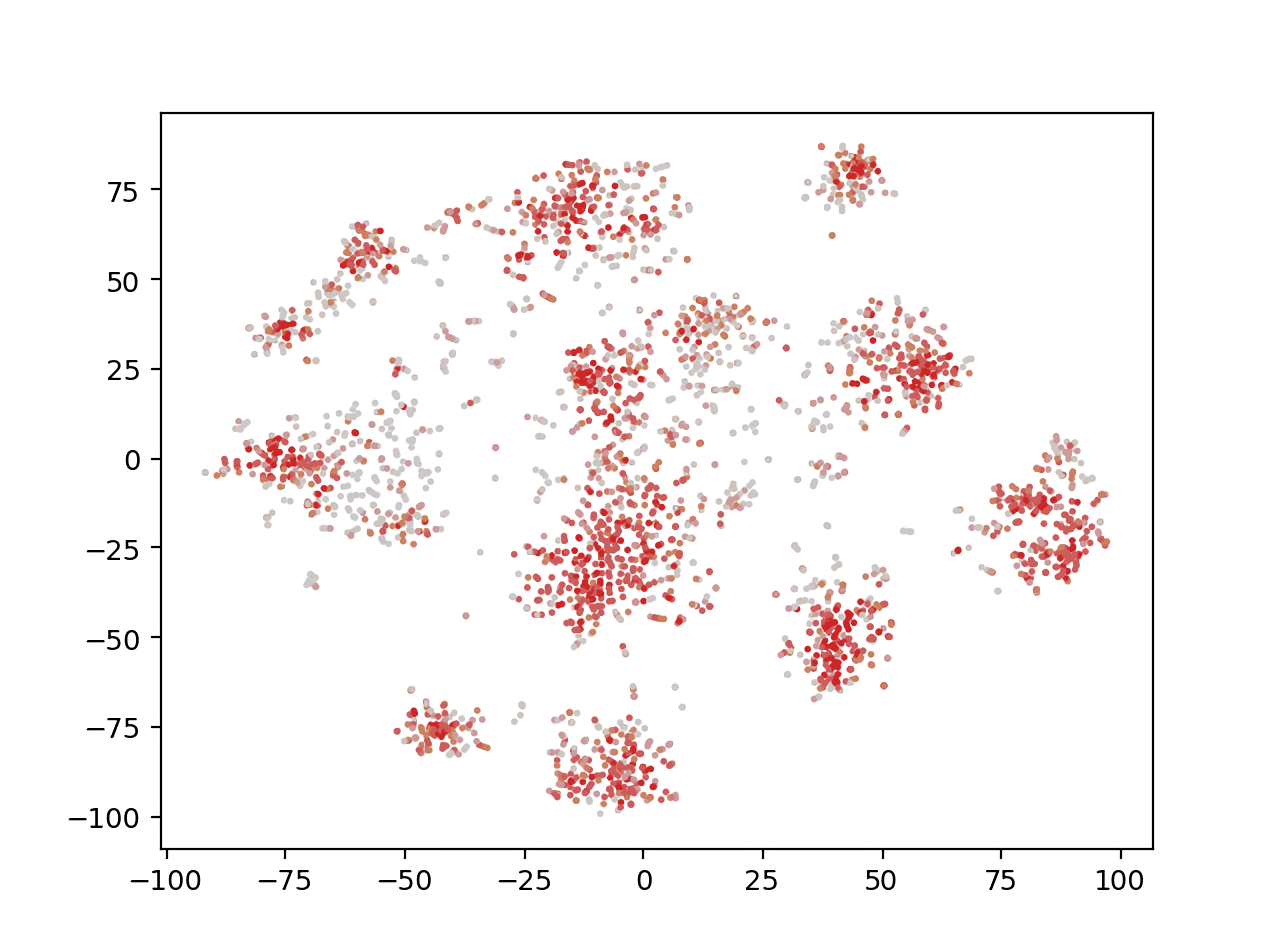}
\caption{Information entropy distribution of deeply learned features in the test set of ScanObjectNN. Darker red corresponds to a lower entropy value and a lower uncertainty of the predicted probability distribution. These samples are distributed mostly in the high-density regions near the centers of the clusters. In contrast, a lighter color indicates that the information entropy value is high; these samples are located mainly in the sparse regions.}
\end{figure}

\section{Experiments}
In the experiments, we employ two benchmark point cloud datasets for evaluation: ModelNet40 [18] and ScanObjectNN [61]. First, we present the experimental settings. Then, we compare the performance of our approach with those of state-of-the-art methods. Furthermore, we analyze the effects of various components of our network by conducting relevant ablation studies.

\subsection{Experimental Settings}
\subsubsection{Implementation}
Our method is a general component that can be feasibly embedded into any point cloud classification network. It can increase the classification performance of the backbone network. For example, we adopt GBNet \cite{ref31} as a backbone. After the backbone network maps the input data into a 2048-dimensional embedded feature, we use our supervised contrastive learning method to refine the embedded feature. Specifically, the CPCM module and the EAA module are utilized in each batch to calculate the contrastive loss. Finally, the contrastive loss and cross-entropy loss are combined in the training process. This project is implemented with PyTorch, and the training and testing processes are carried out on a workstation with 8 GeForce RTX 2080Ti GPUs.

\begin{figure*}
\centering
\includegraphics[width=18cm]{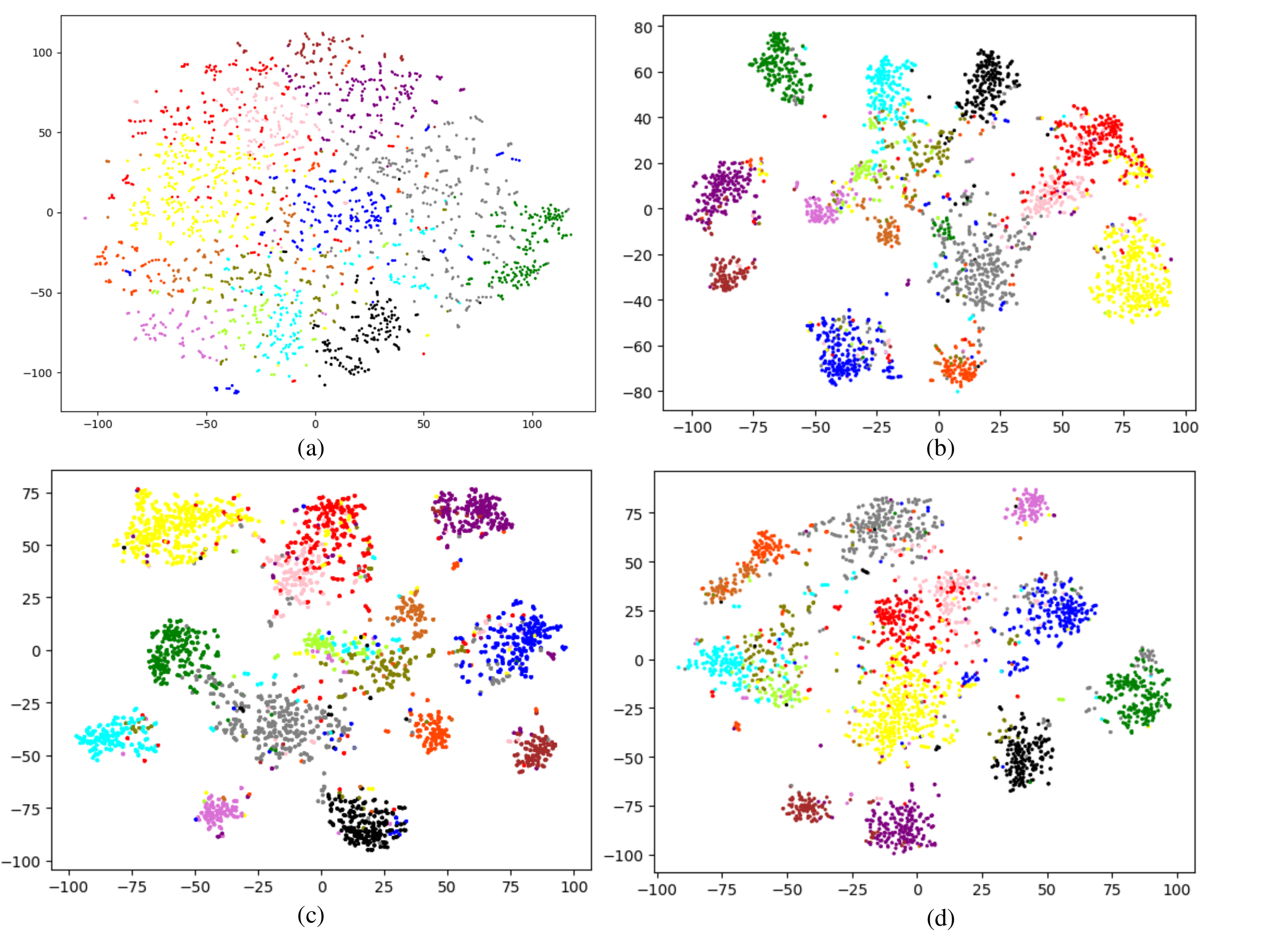}
\caption{T-SNE visualization of point cloud classification for the GBNet backcbone on the ScanObjectNN dataset. (a)  baseline, (b) supervised contrastive learning, (c) 	entropy-aware attention, and (d) our approach that combine CPCM with (C) .}
\end{figure*}

\subsubsection{Training}
We apply stochastic gradient descent (SGD) with a momentum of 0.9 as the optimizer for training, and its learning rate is decreased from 0.1 to 0.001 by cosine annealing \cite{ref62}. The batch size is set to 80 for 300 epochs during training. We also use L2 regularization.

\subsubsection{Datasets}
Our method has been tested on the ModelNet40 and ScanObjectNN benchmarks for comparison with the state-of-the-art methods.

\textbf{ModelNet40} is a synthetic point cloud dataset that was sampled from CAD models, which comprises 9832 training objects and 2468 test objects in 40 classes. Moreover, the corresponding points are uniformly sampled from the mesh surfaces and further preprocessed by moving to the origin and scaling into a unit sphere. As the most widely used benchmark for point cloud classification, ModelNet40 is popular and authoritative because of its clean shapes and well-constructed data. In our experiments, we use the 3D coordinates $(x,y,z)$ of 1024 points for each point cloud sample as input.

\textbf{ScanObjectNN}. To further evaluate the performance and robustness of our approach, we conduct experiments on ScanObjectNN, which is a newly published real-world object dataset with approximately 15k objects in 15 categories. Although it has fewer categories than ModelNet40, it is more practically challenging than its synthetic counterpart due to complex backgrounds, missing parts, and various deformations.

\begin{table}[htbp]
  \centering
  \caption{Classification result (\%) on the ModelNet40 benchmark.(COORDS:$(x,y,z)$ coordinates, NORM: point normal, VOTING: multi-votes evalution strategy, and-:unknown)}
    \begin{tabular}{cccc}
    \toprule
    \toprule
    Method & Input & Avg class acc & Overall acc \\
    \midrule
    3DmfV \cite{ref63} & \textit{coords} & 86.3  & 91.4 \\
    PointNet \cite{ref20} & \textit{coords} & 86.2  & 89.2 \\
    PointNet++ \cite{ref21} & \textit{coords} & 87.8  & 90.7 \\
    SRN-PN++ \cite{ref23} & \textit{coords} & -     & 91.5 \\
    PointCNN \cite{ref26} & \textit{coords} & 88.1  & 92.2 \\
    pointWeb \cite{ref22} & \textit{coords} & 89.4  & 92.3 \\
    PointConv \cite{ref78} & \textit{coords} & -     & 92.5 \\
    DGCNN \cite{ref29} & \textit{coords} & 90.2  & 92.2 \\
    RS-CNN \cite{ref25} & \textit{coords} & -     & 92.9 \\
    \midrule
    Ours w/PN++ & \textit{coords} & \textbf{91.1} & \textbf{93.1} \\
    Ours w/DGCNN & \textit{coords} & \textbf{89.7} & \textbf{93.2} \\
    \midrule
    PointNet++ \cite{ref21} & \textit{coords + normal} & -     & 91.9 \\
    \bottomrule
    \bottomrule
    \end{tabular}%
  \label{tab:addlabel}%
\end{table}%

In the evaluation, we used the most complex $PB_T75_RS$ split, which extends the original object by translation, rotation, and scaling. Suffix T75 denotes translation that randomly shifts the bounding box by up to 75\% of its size from the box centroid along each world axis. Suffixes R and S denote rotation and scaling. In all experiments, we uniformly sample 1024 points for each point cloud for training and evaluation, and all our results are measured using a single view without the multiview voting trick to show the performances of the compared models.

\subsection{Classification Performance}
\subsubsection{Comparison on Synthetic ModelNet40}
As presented in Table \uppercase\expandafter{\romannumeral1}, quantitative results are obtained on the synthetic ModelNet40 classification benchmark. Our approach with the PointNet backbone achieves a competitive test accuracy of 93.1\% compared to the original PointNet++ \cite{ref21} (+2.4 \%). The DGCNN backbone achieves a competitive test accuracy of 93.1\% compared to the original DGCNN \cite{ref29} (+2.4 \%). In addition, the boosting of our method is even more effective than increasing the input of the outer normal vector (91.9\%).

These well-constructed point cloud samples are clear and complete in shape and free of noise. Consequently, the small intra-class and large inter-class variations within the datasets render them easy to classify. For such data, some encoders with sufficient performance can learn features with good separability and discrimination. The advantage of our model is that it makes the deeply learned feature more compact. As a result, our approach limits enhancements to some established high-performance models and achieves a larger boost in PointNet++ \cite{ref21} (2.4\%) than in DGCNN \cite{ref29} (1\%). This result also shows that our method can compensate for the lack of backbone performance and improve the classification performance to the same level.

\begin{table*}[htbp]
  \centering
  \caption{Classification results (\%) on the ScanObjectNN  benchmark  (percision in  each  class)}
    \begin{tabular}{cccccccccccccccc}
    \toprule
    \toprule
    Methods & Bag   & Bin   & Box   & Cabinet & Chair  & Desk  & Display & Door  & Shelf  & Table & Bed   & Pillow & Sink  & Sofa  & Toilet \\
    \midrule
    3DmfV & 39.8  & 62.8  & 15    & 65.1  & 84.4  & 36    & 62.3  & 85.2  & 60.6  & 66.7  & 51.8  & 61.9  & 46.7  & 72.4  & 61.2 \\
    PointNet & 36.1  & 69.8  & 10.5  & 62.6  & 89    & 50    & 73    & \textbf{93.8} & 72.6  & 67.8  & 61.8  & 67.6  & 64.2  & 76.7  & 55.3 \\
    PointNet++ & 49.4  & 84.4  & 31.6  & 77.4  & 91.3  & 74    & 79.4  & 85.2  & 72.6  & 72.6  & 75.5  & \textbf{81} & \textbf{80.8} & 90.5  & \textbf{85.9} \\
    PointCNN & 57.8  & 82.9  & 33.1  & 83.6  & 92.6  & 65.3  & 78.4  & 84.8  & 84.2  & 67.4  & 80    & 80    & 72.5  & 91.9  & 71.8 \\
    DGCNN & 49.4  & 82.4  & 33.1  & 83.9  & 91.8  & 63.3  & 77    & 89    & 79.3  & \textbf{77.4} & 64.5  & 77.1  & 75    & 91.4  & 69.4 \\
    BGA-PN++ & 54.2  & \textbf{85.9} & 39.8  & 81.7  & 90.8  & 76    & 84.3  & 87.6  & 78.4  & 74.4  & 73.6  & 80    & 77.5  & 91.9  & 85.9 \\
    BGA-DGCNN & 48.2  & 81.9  & 30.1  & \textbf{84.4} & 92.6  & \textbf{77.3} & 80.4  & 92.4  & 80.5  & 74.1  & 73.6  & 80    & 77.5  & 91.9  & \textbf{85.9} \\
    GBNet & 59    & 84.4  & 44.4  & 78.2  & 92.1  & 66    & 91.2  & 91    & 86.7  & 70.4  & \textbf{82.7} & 78.1  & 72.5  & \textbf{92.4} & 77.6 \\
    \midrule
    Ours w/PN++ & 48.2  & 80.9  & 52.6  & 82.5  & 92.3  & 68    & 87.3  & 91.4  & 85.5  & 72.6  & \textbf{82.7} & 71.4  & 77.5  & 92    & 80 \\
    Ours w/DGCNN & 57.8  & 87.9  & 54.9  & 81.2  & 93.1  & 57.3  & 88.7  & 90.5  & 80.5  & 73    & 77.3  & \textbf{81} & 78.3  & \textbf{92.4} & 72.9 \\
    Ours w/GBNet & \textbf{63.9} & 85.43 & \textbf{63.2} & 76.9  & \textbf{93.4} & 68.7  & \textbf{95.6} & 90    & \textbf{88} & 74.1  & 80.9  & 79    & \textbf{80.8} & 90    & 78.8 \\
    \bottomrule
    \bottomrule
    \end{tabular}%
  \label{tab:addlabel}%
\end{table*}%

\begin{table}[htbp]
  \centering
  \caption{Classification result (\%) on the ScanObjectNN benchmark.(coords: $(x,y,z)$ coordinates, norm: point normal, voting: multi-votes evalution strategy, and-:unknown)}
    \begin{tabular}{ccccc}
    \toprule
    \toprule
    Method & Input & Overall  & Avg class acc. & Overall acc. \\
          &       &       &       &  \\
    \midrule
    3DmfV & \textit{coords} & -     & 58.1  & 63 \\
    PointNet & \textit{coords} & -     & 63.4  & 68.2 \\
    PointNet++ & \textit{coords} & -     & 75.4  & 77.9 \\
    PointCNN & \textit{coords} & 0.78  & 75.1  & 78.5 \\
    DGCNN & \textit{coords} & 0.78  & 73.6  & 78.1 \\
    BGA-PN++ & \textit{coords} & -     & 77.5  & 80.2 \\
    BGA-DGCNN & \textit{coords} & -     & 75.7  & 79.7 \\
    GBNet & \textit{coords} & 0.8   & 77.8  & 80.5 \\
    \midrule
    Oursw/PointNet++ & \textit{coords} & \textbf{0.81} & \textbf{77.7} & \textbf{81} \\
    Ours w/DGCNN & \textit{coords} & \textbf{0.81} & \textbf{77.9} & \textbf{80.8} \\
    Ours w/GBNet & \textit{coords} & \textbf{0.83} & \textbf{80.6} & \textbf{82.9} \\
    \midrule
    MVTN  & \textit{+voting} & -     & -     & 82.8 \\
    \bottomrule
    \bottomrule
    \end{tabular}%
  \label{tab:addlabel}%
\end{table}%

\begin{figure}
\centering
\includegraphics[width=9cm]{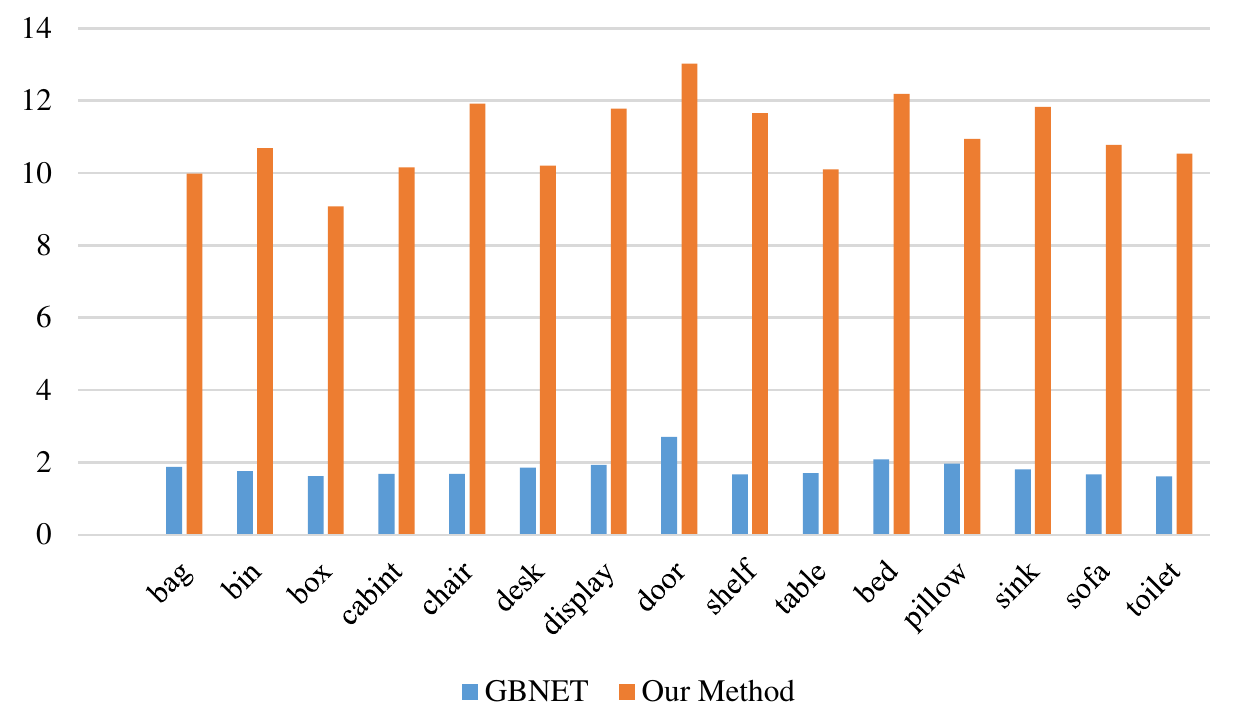}
\caption{Our approach based on the GBNet backbone vs. the GBNet baseline. Sum of the distances between each cluster center and the centers of all other classes of clusters. A larger sum of distances between classes after our method is applied indicates that the higher-level features are more widely separated.}
\end{figure}

\subsubsection{Comparison on Realistic ScanObjectNN}
For the real-world 3D point cloud dataset, we use the same architecture and training strategy to evaluate our method. As presented in TABLE \uppercase\expandafter{\romannumeral3}, additionally considering the cross-instance representation contrast in classification network learning leads to substantial performance gains in several backbones, namely, 2.9\% in DCGNN, 3.1\% in PointNet++ and 2.4\% in GBNet. The results of our network significantly improve the classification accuracy, with an overall accuracy of 82.9\% and an average class accuracy of 80.6\% on the benchmark, thereby highlighting the advantages of our approach for realistic 3D point cloud scans. Furthermore, our approach performs even better than MVTN, which uses a multiview network with an extra input of 12 view images. In terms of the average accuracy of each class (\uppercase\expandafter{\romannumeral2}), our method outperforms the other methods in 7 out of 15 categories; for the most challenging class, namely, box, the accuracy is improved from 44.4\% to 63.2\% with the same GBNet backbone.

The ScanObjectNN dataset contains many challenging cases. One problem is inadequate structural information, where only a few points provide semantic information due to low reconstruction accuracy or occlusion. In this case, it is difficult to extract the semantic feature. However, our approach with the SCC module enables the network to mine similarities from more similar samples, thereby compensating for the problem of inadequate structural information of a single sample. Another problem is that the remaining background points in real-world data may confuse the network because they are irrelevant to the shape structure of the target category. Our entropy-aware attention module attempts to reweight the most background-sensitive samples according to their entropy, by which the importance of the background can be reduced. Moreover, the most challenging problem of real-world datasets is that some objects share similar geometric features (\emph{e.g.}, tables and desks share the table plane). Therefore, these categories can easily be confused with each other. To avoid confusion, we not only use the supervised contrastive classification module to extract more discriminatory features but also design the online confusing class mining module to separate the most easily confused class pairs.

To further evaluate the performance and robustness, we also examine the $F_{1}$ score, which is the harmonic mean of the precision and recall, as another quantitative measurement. We compare our method in terms of $F_{1}$ score with the best two state-of-the-art methods on the ScanObjectNN official leaderboard \cite{ref72}, namely, GBNet \cite{ref31} and DGCNN \cite{ref29}, along with the BGA-based methods in \cite{ref12}. As presented in Table \uppercase\expandafter{\romannumeral3}, our approach achieves better results in terms of both overall and average class F1 scores. In general, our network has a better balance between the metric precision and recall for real-world point cloud classification.

\begin{table}[htbp]
  \centering
  \caption{Ablation analysis of Our approach on the ScanObjectNN with the GbNet Backbone.}
    \begin{tabular}{ccccc}
    \toprule
    \toprule
    SCC & CPCM & EAA & Avg class acc & Overall acc. \\
    \midrule
    -     & -     & -     & 77.8  & 80.5 \\
    \checkmark     & -     & -     & 78.4  & 81.4 \\
    \checkmark     & \checkmark     & -     & 79.8  & 82.3 \\
    \checkmark     & -     & \checkmark     & 80.1  & 82.4 \\
    \checkmark     & \checkmark     & \checkmark     & 80.6  & 82.9 \\
    \bottomrule
    \bottomrule
    \end{tabular}%
  \label{tab:addlabel}%
\end{table}%

\begin{table}[htbp]
  \centering
  \caption{Ablation analysis of Our approach on the ModelNet40 with the PointNet++ Backbone.}
    \begin{tabular}{ccccc}
    \toprule
    \toprule
    SCC & CPCM & EAA & Avg class acc & Overall acc. \\
    \midrule
    -     & -     & -     & 87.8  & 90.7 \\
    \checkmark     & -     & -     & 88.4  & 91.7 \\
    \checkmark     & \checkmark     & -     & 89.6  & 92.5 \\
    \checkmark     & -     & \checkmark     & 90.0  & 92.6 \\
    \checkmark     & \checkmark     & \checkmark     & 91.1  & 93.1 \\
    \bottomrule
    \bottomrule
    \end{tabular}%
  \label{tab:addlabel}%
\end{table}%

\subsection{Ablation Study}
To verify the functions and effectiveness of components in our network, we conduct ablation studies on the modules that are proposed in this work. In addition, we examine the network complexity and visualize the learned features.

\subsubsection{Supervised Contrastive Classification}
First, we investigate the effectiveness of the SCC module. As presented in Table \uppercase\expandafter{\romannumeral4}, considering cross-instance semantic relations in the ScanObjectNN datasets leads to a substantial performance gain (overall acc.: 0.9\% and avg. class acc.: 0.6\%) compared with ``baseline (w/o contrast)''. This demonstrates that enhanced intra-cluster compactness and inter-cluster separability using contrastive learning can lead to better classification performance. We also conduct an ablation study on ModelNet40, and the performance of SCC reaches 1\% in overall acc. and 0.6\% in avg. class acc., as presented in Table \uppercase\expandafter{\romannumeral5}.

In addition, we examine various values of the balance coefficient $\lambda$ in Eq. 3 for joint training, and the results are presented in Table \uppercase\expandafter{\romannumeral6}. We implement fixed and variable coefficients, where the fixed coefficients remain constant throughout the training process and the variable coefficients are increased with the number of iterations (gradually increased from 0.1 to 0.2). The results show that fixed coefficients outperform variable coefficients. The value of this hyperparameter has a significant impact on the results. Of the examined values, 0.1 is the most suitable;

\begin{table}[htbp]
  \centering
  \caption{Values of coefficient $\lambda$ with the supervised contrastive classification module on the ScanObjectNN with the GBNet Backbone.}
    \begin{tabular}{ccc}
    \toprule
    \toprule
    Constant vs Variable &   $\lambda$    & Overall acc. \\
    \midrule
    Constant & 0.3   & 80.1 \\
    Constant & 0.2   & 80.7 \\
    Variable & 0.1$\sim$0.2 & 79.8 \\
    Constant & 0.1   & 81.4 \\
    Constant & 0.05  & 80.9 \\
    \bottomrule
    \bottomrule
    \end{tabular}%
  \label{tab:addlabel}%
\end{table}%

\subsubsection{Confusion-prone Classes Mining}
To resolve the confusion in the classification task, we design a contrastive learning loss based on the distances of cluster centers, which is named the confusion-prone classes mining module. As presented in Tables \uppercase\expandafter{\romannumeral4} and \uppercase\expandafter{\romannumeral5}, adding this module can improve the performance by 0.9\% on ScanobjectNN and 0.8\% on ModelNet40 compared to using only the supervised contrastive classification module. The reasoning behind this design is that the degree of confusion is highly correlated with the distance between feature cluster centers. We find that the closer the category cluster centers are to each other, the more likely they are to be confused, as shown by the confusion matrix in Fig. 8 and the corresponding cluster center distance matrix in fig. 9. As shown in Fig. 10, after our CPCM method, the clusters are well separated, and the sum of the distances between the cluster centers increases.

\begin{figure}
\centering
\includegraphics[width=9cm]{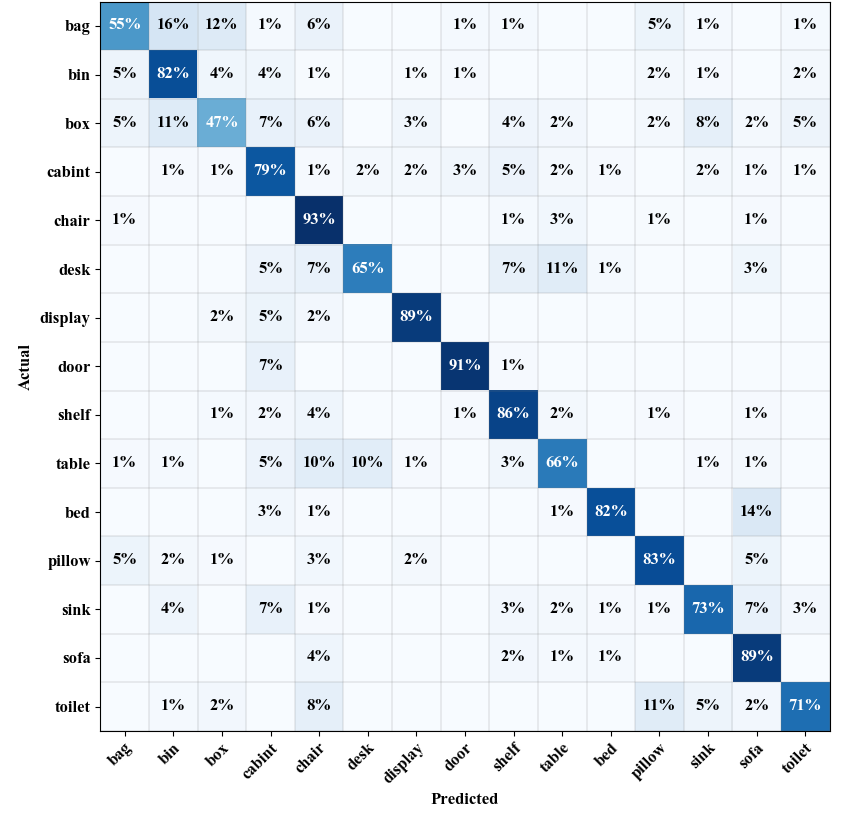}
\caption{Confusion matrix of our approach in ScanObjectNN.}
\end{figure}

\begin{figure}
\centering
\includegraphics[width=9cm]{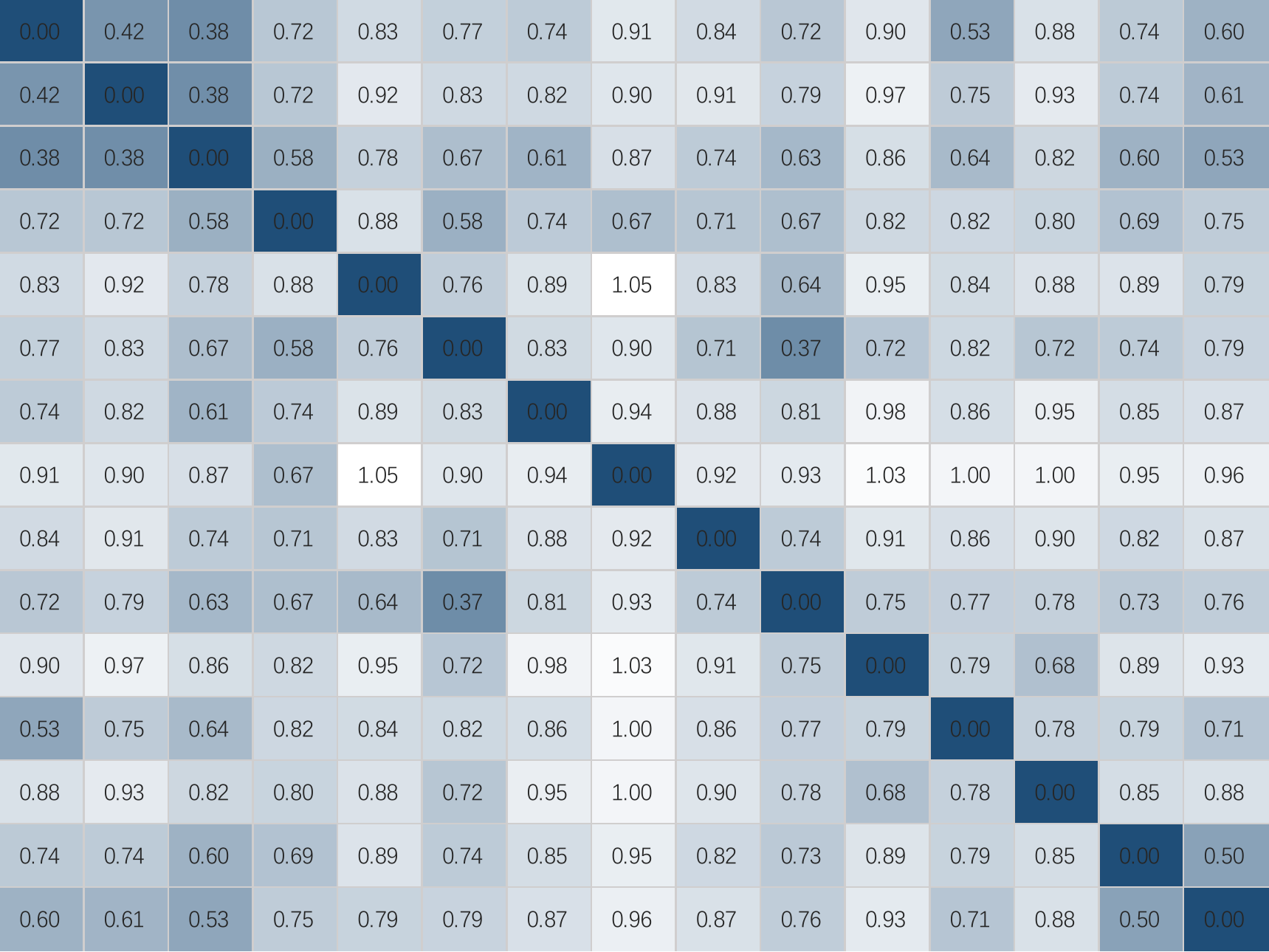}
\caption{Embedding feature cluster center distance between each class pairs in ScanObjectNN. Combined with Figure 9, it can be seen that the smaller the cluster center distance, the easier it is to confuse.}
\end{figure}

We also implement two confusion-prone classes mining methods. In Method 1, as introduced Section \uppercase\expandafter{\romannumeral3} (Eq. 5.), all pairs with other categories are considered for each class. We designed Method 2 to follow the most recent category only. For each category, we mined the category pair with the smallest distance by a large margin (0.8, which is an experience value) than the second smallest category pair. As presented in TABLE \uppercase\expandafter{\romannumeral7}, Method 1 is better than method 2, and both prove the validity of the module. A possible explanation for this might be that confusion usually occurs in multiple categories instead of a single class.

\begin{table}[htbp]
  \centering
  \caption{Various Methods with Our confusion-prone classes mining Module on the ScanObjectNN with the GBNet Backbone.}
    \begin{tabular}{rccc}
    \toprule
    \toprule
    \multicolumn{1}{c}{SCC} & CPCM  & Avg. class Acc. & Overall Acc. \\
    \midrule
      \checkmark   & Method1 & 79.8  & 82.3 \\
      \checkmark   & Method2 & 79.6  & 82.1 \\
    \bottomrule
    \bottomrule
    \end{tabular}%
  \label{tab:addlabel}%
\end{table}%

\subsubsection{Entropy-Aware Attention}
The entropy-aware attention module is another vital component for improving the overall performance of the model. As presented in Table \uppercase\expandafter{\romannumeral4}, this module can improve the performance of the model by 0.9\%. On ModelNet40 (as presented in Table V), this module can also improve the performance of the model by 0.9\%. Predicting the classification result is a process in which the classifier finds a hyperplane in the feature space. The classifier makes predictions based on the position relationship between the hyperplane and samples. Some samples near the decision boundary have a significant impact on the model. In addition, some outliers that are close to the cluster centers but misclassified are also significant factors that affect the performance. Based on this observation, we design ECC to reweight the importance of these particular samples.

We design reweighting mechanisms to explore the effectiveness of the module. The varying weight is introduced in Section \uppercase\expandafter{\romannumeral3} (Eq. 7). We propose a fixed weight for outlier and unstable nodes in the contrastive learning loss, which is defined as follows:
\begin{equation}
a_{P}=\left\{\begin{array}{cc}
0.8 ; & p \subset \text { outliers } \\
1.2 ; & p \subset \text { Unstable Nodes }
\end{array}\right.
\end{equation}

In Table \uppercase\expandafter{\romannumeral8}, the varying weight outperforms the fixed weight by 0.5\%. One possible reason is that the disparate treatment of samples with different entropy values exerts a tremendous influence.

\begin{table}[htbp]
  \centering
  \caption{Various Methods with Our Entropy-aware Attention Module on the ScanObjectNN with the GBNet Backbone.}
    \begin{tabular}{rccc}
    \toprule
    \toprule
    \multicolumn{1}{c}{SCC} & ECC  & Avg. class Acc. & Overall Acc. \\
    \midrule
      \checkmark   & Fixed weight & 79.6  & 81.9 \\
      \checkmark   & Varying weight & 80.1  & 82.4 \\
    \bottomrule
    \bottomrule
    \end{tabular}%
  \label{tab:addlabel}%
\end{table}%

\subsubsection{Feature Visualization}
In Fig. 7, the corresponding learned features are visualized by t-distributed stochastic neighbor embedding (t-SNE) \cite{ref72}. The t-SNE method maps an embedding from high-dimensional space into low-dimensional space by minimizing the differences in all pairwise similarities between points in high- and low-dimensional spaces.

As shown in Fig. 7(b), the learned embedding by supervised contrastive classification becomes much more compact and well separated compared to Fig. 7(a). This suggests that our SCC approach can generate more discriminative features by utilizing embedding feature refinement, thereby producing better representation performance.

However, some hard cases clusters( \emph{e.g.} green-yellow, olive, and aqua) remain in a heavily mixed situation, which will make it difficult to search the optimal classification boundary for these categories. Besides, some green nodes representing the door appear in the gray clusters, indicating that these samples are more likely to be misclassified as cabinets (in gray color). After dealing with unstable points by the EAA module, almost all clusters become more clearly bounded (Fig. 7(c)). For example, the green nodes become well clustered and are not mixed with the gray nodes, presumably a credit to the efficient processing of outliers.

As shown in (Fig. 7(b)), after combining CPCM module, we observe that the previously challenging confusion-prone classes (\emph{e.g.} red vs. pink, which represent desk and table) are basically well clustered and easily separated. This demonstrates that our strategies lead to more robust decision boundaries and better embedding distributions for most categories.

\section{Conclusion and Discussion}
In this paper, we propose a new point-based network in a self-supervised contrastive learning fashion targeting point cloud classification task. To leverage more global semantic relationships between individual samples, we explicitly propose a supervised contrastive classification (SCC) method by a joint training method with the contrastive loss and cross-entropy loss. Accordingly, the confusion-prone classes mining (CPCM) module avoids confusion by mining classes with small inter-class variations. In addition, the entropy-aware attention (EAA) module further refines the learned embedding feature by focusing on hard cases. Compared with some state-of-the-art methods, our method achieves salient improvement on classification task of synthetic and real-world point cloud datasets. Besides, we validate the properties of the proposed modules through necessary ablation studies and visualizations. The results show the effectiveness and robustness of our approach.

Actually, this paper provides a generic solution to the classification task, which is not only towards the point cloud data, but also other data patterns. Moreover, it faces fundamental issues as confusion and overfitting in the classification topic. More tasks, \emph{e.g.} object detection, instance segmentation, will be explored in future.

\ifCLASSOPTIONcaptionsoff
  \newpage
\fi

%Formatter's Note: Your references and in-text citations have been formatted to conform to the journal's guidelines. We encourage you to keep these changes.


\begin{IEEEbiography}[{\includegraphics[width=1in,height=1.25in,clip,keepaspectratio]{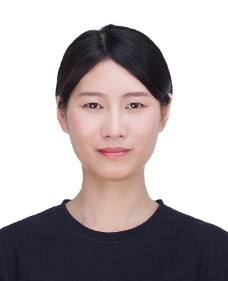}}]{Feng Yang}
received the M.S. degree in School of Instrument Science and Engineering, Southeast University, Nanjing, China, in 2020. Where she is currently pursuing a Ph.D. degree.  Her research interests include 3D scene understanding, deep learning on point cloud , simultaneous localization and mapping.
\end{IEEEbiography}
%\vspace{-10 mm}
\begin{IEEEbiography}[{\includegraphics[width=1in,height=1.25in,clip,keepaspectratio]{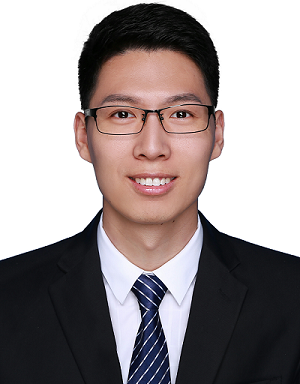}}]{Yichao Cao}
received the M.S. degree in School of Automation from Southeast University, Jiangsu, China, in 2019. Where is currently pursuing the Ph.D. degree. He is currently an algorithm researcher with Nanjing Enbo Technology Co., Ltd. He is doing his doctoral research with Xiaobo Lu Laboratory and his research interests include image processing, deep learning and computer vision. He is a Reviewer for international journals, such as IEEE Transactions on Multimedia, Neural Networks, and IEEE Transactions on Cognitive and Developmental Systems.
\end{IEEEbiography}
%\vspace{-10 mm}
\begin{IEEEbiography}[{\includegraphics[width=1in,height=1.25in,clip,keepaspectratio]{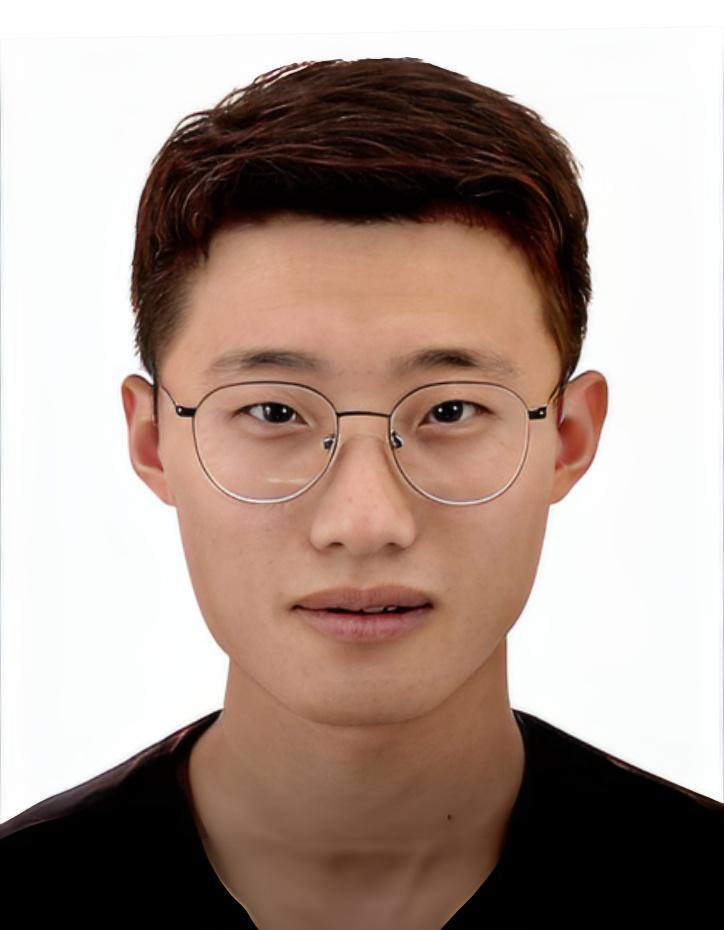}}]{Qifan Xue}
received his B.S. degree in School of Instrument Science and Engineering from Southeast University in 2017. He is currently a Ph.D. student in School of Instrument Science and Engineering, Southeast University, China. His research interests include causal perception and scene understanding on the traffic scene.
\end{IEEEbiography}
%\vspace{-10 mm}
\begin{IEEEbiography}[{\includegraphics[width=1in,height=1.25in,clip,keepaspectratio]{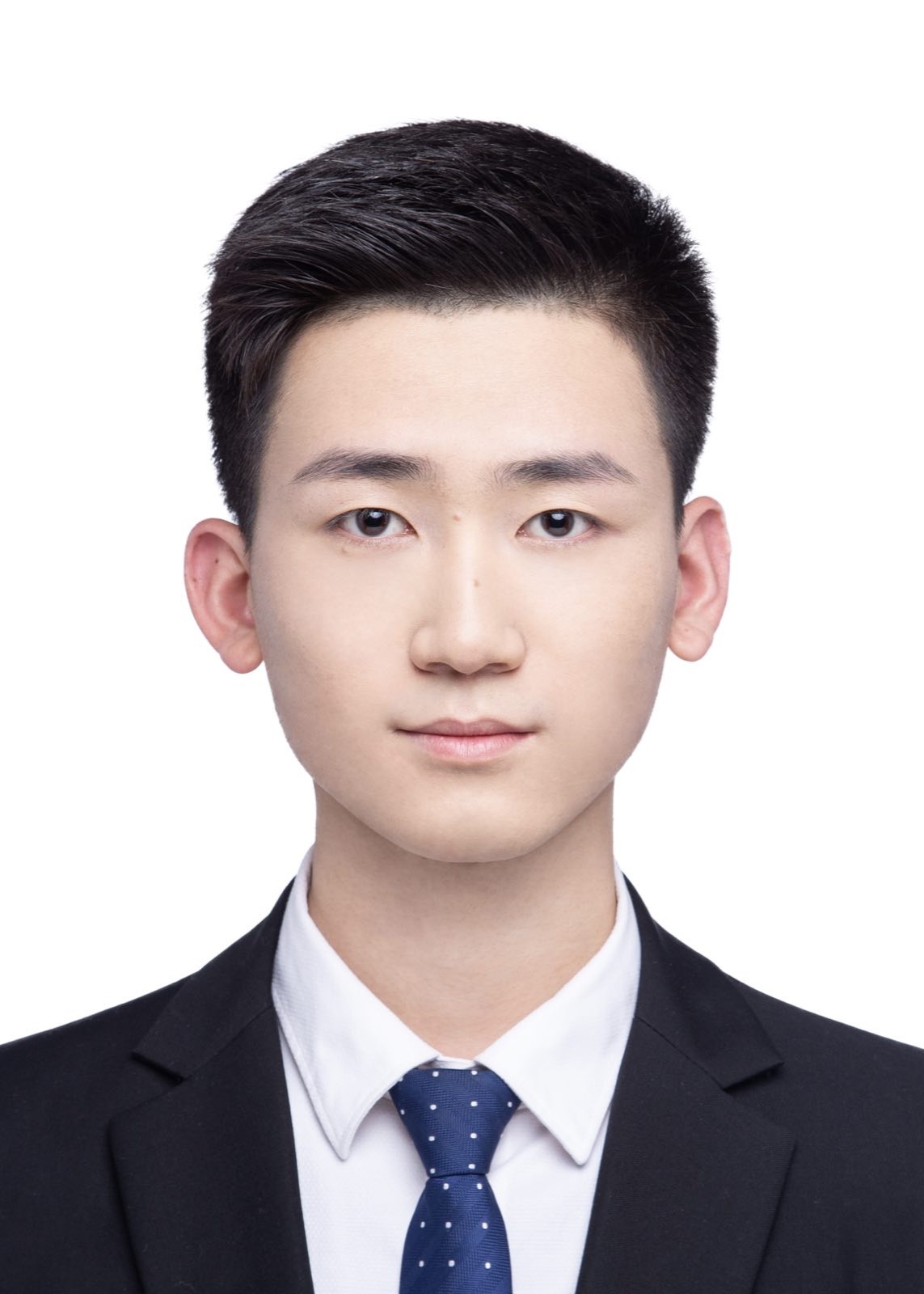}}]{Shuai Jin}
received his B.S. degree in School of Instrument Science and Engineering from Southeast University in 2020. He is currently a master student in School of Instrument Science and Engineering, Southeast University, China. His research interests include 3D scene understanding, deep learning on point cloud and object detection.
\end{IEEEbiography}
%\vspace{-140 mm}
\begin{IEEEbiography}[{\includegraphics[width=1in,height=1.25in,clip,keepaspectratio]{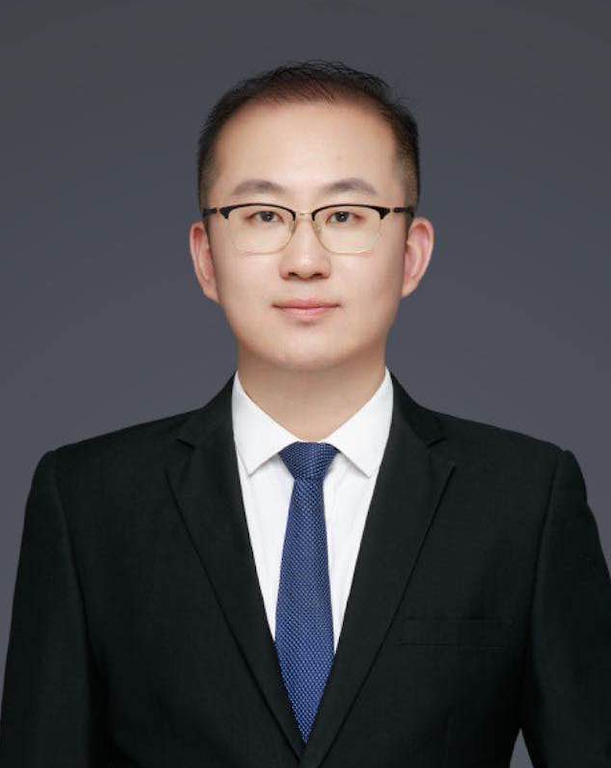}}]{Xuanpeng Li}
received his B.S. and M.S. degrees in instrument science and technology from Southeast University, China, in 2007 and 2010, and his Ph.D. degree in information technology from the Universit e de Technologie de Compi`egne, France, in 2014. From 2014 to 2015, he was a post-doctor at LIVIC, IFSTTAR, in France. Since 2021, he has been an associate professor with the School of Instrument Science and Engineering. His research interests include 3D scene understanding, causal perception, and risk estimation for intelligent transportation systems.
\end{IEEEbiography}
%\vspace{-140 mm}
\begin{IEEEbiography}[{\includegraphics[width=1in,height=1.25in,clip,keepaspectratio]{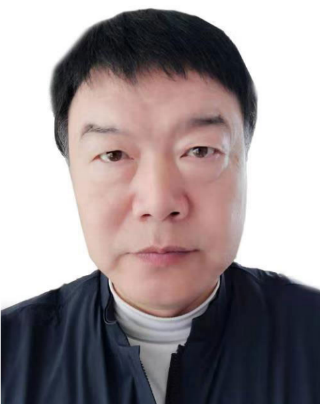}}]{Weigong Zhang}
received the B.S. degree from the Faculty of Mathematics and Mechanics, Nanjing Aviation Academy, Nanjing, China, the M.S. degree in solid mechanics from Nanjing Aeronautics and Astronautics University, Nanjing, and the Ph.D. degree in precision instruments and mechanical engineering from Southeast University, Nanjing. He is currently a Professor at Southeast University. His research interests include automotive electronics and mechatronics.
\end{IEEEbiography}


\begin{thebibliography}{99}

\bibitem{ref1}	D. Maturana and S. Scherer, ``Voxnet: A 3d convolutional neural network for real-time object recognition,'' in 2015 \emph{IEEE/RSJ International Conference on Intelligent Robots and Systems (IROS)}. IEEE, 2015, pp.
922–928.
\bibitem{ref2}	S. Song, F. Yu, A. Zeng, A. X. Chang, M. Savva, and T. Funkhouser, ``Semantic scene completion from a single depth image,'' in \emph{Proceedings of the IEEE Conference on Computer Vision and Pattern Recognition}, 2017, pp. 1746-1754.
\bibitem{ref3}	A. Oord, D. V, Y. Li, and O. Vinyals, ``Representation learning with contrastive predictive coding.'' \emph{arXiv preprint arXiv}:1807.03748, 2018.
\bibitem{ref4}	R. D. Hjelm, A. Fedorov,  S. Lavoie-Marchildon, K. Grewal, P. Bachman,  A. Trischler, and Y.Bengio, ``Learning deep representations by mutual information estimation and maximization.'' \emph{arXiv preprint arXiv}:1808.0667, 2018.
\bibitem{ref5}	K. He, X. Zhang, S. Ren, and J. Sun, ``Deep residual learning for image recognition.'' In \emph{Proceedings of the IEEE conference on computer vision and pattern recognition}. 2016, pp. 770-778.
\bibitem{ref6}	J. B. Grill, F.  Strub, , F Altché, C. Tallec, P. H. Richemond, and E. Buchatskaya, ``Bootstrap your own latent: a new approach to self-supervised learning.'' \emph{arXiv preprint arXiv}:2006.07733, 2020.
\bibitem{ref7}	X. Chen, and K. He, ``Exploring simple siamese representation learning.'' In \emph{Proceedings of the IEEE/CVF Conference on Computer Vision and Pattern Recognition}, 2021, pp. 15750-15758.
\bibitem{ref8}	M. Caron, P. Bojanowski, A. Joulin, M. Douze, ``Deep clustering for unsupervised learning of visual features.'' In \emph{Proceedings of the European Conference on Computer Vision (ECCV)}, 2018, pp. 132-149.
\bibitem{ref9}	M. Caron, I. Misra, J. Mairal, P. Goyal, P. Bojanowski, and A. Joulin, ``Unsupervised learning of visual features by contrasting cluster assignments.'' \emph{arXiv preprint arXiv}:2006.09882, 2020.
\bibitem{ref10}	T. Chen,  S. Kornblith, M. Norouzi, and G. Hinton, ``A simple framework for contrastive learning of visual representations.'' In \emph{International conference on machine learning}. PMLR, 2020, pp. 1597-1607.
\bibitem{ref11}	H. Lu, and H. Shi. ``Deep Learning for 3D Point Cloud Understanding: A Survey.'' \emph{arXiv preprint arXiv}:2009.08920, 2020.
\bibitem{ref12}	H. Su, S. Maji ,E. Kalogerakis, and E. Learned-Miller, ``Multi-view convolutional neural networks for 3d shape recognition.'' In \emph{Proceedings of the IEEE international conference on computer vision}, 2015, pp. 945-953.
\bibitem{ref13}	Y. Feng, Z. Zhang, X. Zhao, R Ji, and Y. Gao, ``GVCNN: Group-view convolutional neural networks for 3D shape recognition.''  In \emph{Proceedings of the IEEE Conference on Computer Vision and Pattern Recognition}, 2018, pp. 264–272.
\bibitem{ref14}	T. Yu, J. Meng, and J. Yuan, ``Multi-view harmonized bilinear network for 3d object recognition.'' In \emph{Proceedings of the IEEE Conference on Computer Vision and Pattern Recognition}, 2018, pp. 186-194.
\bibitem{ref15}	Z. Yang, and L. Wang. ``Learning relationships for multiview 3D object recognition.'' In \emph{Proceedings of the IEEE International Conference on Computer Vision}, 2019, pp 7505–7514.
\bibitem{ref16}	A. Hamdi, S. Giancola, and B. Ghanem, ``MVTN: Multi-View Transformation Network for 3D Shape Recognition.'' In \emph{Proceedings of the IEEE/CVF International Conference on Computer Vision}, 2021, pp. 1-11.
\bibitem{ref17}	A. X. Chang, T. Funkhouser, L. Guibas, P. Hanrahan, Q. Huang, Z. Li, S. Savarese, M. Savva, S. Song, H. Su, J. Xiao, L. Yi and Yu, F, ``Shapenet: An information-rich 3d model repository.'' \emph{arXiv preprint arXiv}:1512.03012, 2015.
\bibitem{ref18}	Z. Wu, S. Song, A. Khosla, F. Yu, L. Zhang, X. Tang, and J. Xiao, ``3d shapenets: A deep representation for volumetric shapes,'' in \emph{Proceedings of the IEEE conference on computer vision and pattern recognition}, 2015, pp. 1912–1920.
\bibitem{ref19}	P. S. Wang, Y. Liu, Y. X. Guo, C.Y. Sun, and X. Tong, ``O-CNN: Octree-based convolutional neural networks for 3D shape analysis.'' In \emph{ACM Transactions on Graphics (TOG)}, 2017, vol. 36, no. 4, pp. 1–11.
\bibitem{ref20}	C. R. Qi, H. Su, K. Mo, and L. J. Guibas, ``Pointnet: Deep learning on point sets for 3d classification and segmentation,'' in \emph{Proceedings of the IEEE Conference on Computer Vision and Pattern Recognition}, 2017, pp. 652–660.
\bibitem{ref21}	C. R. Qi, L. Yi, H. Su, and L. J. Guibas, ``Pointnet++: Deep hierarchical feature learning on point sets in a metric space,'' in \emph{Advances in neural information processing systems}, 2017, pp. 5099–5108.
\bibitem{ref22}	H. Zhao, L. Jiang, C.W. Fu, and J. Jia. ``PointWeb: Enhancing local neighborhood features for point cloud processing.'' In \emph{Proceedings of the IEEE Conference on Computer Vision and Pattern Recognition}, 2019, pp. 5565–5573.
\bibitem{ref23}	Y. Duan, Y. Zheng, J. Lu, J. Zhou, and Q. Tian, ``Structural relational reasoning of point clouds.'' In \emph{Proceedings of the IEEE Conference on Computer Vision and Pattern Recognition}, 2019, pages 949–958.
\bibitem{ref24}	A. Komarichev, Z. Zhong, and J. Hua. ``A-CNN: Annularly convolutional neural networks on point clouds.'' In \emph{Proceedings of the IEEE Conference on Computer Vision and Pattern Recognition}, 2019, pages 7421–7430.
\bibitem{ref25}	Y. Liu, B. Fan, S. Xiang, and C. Pan. ``Relation-shape convolutional neural network for point cloud analysis.'' In \emph{Proceedings of the IEEE Conference on Computer Vision and Pattern Recognition}, 2019, pages 8895–8904.
\bibitem{ref26}	Y. Li, R. Bu, M. Sun, W. Wu, X. Di, and B. Chen, ``Pointcnn: Convolution on x-transformed points,'' in \emph{Advances in Neural Information
Processing Systems}, 2018, pp. 820–830.
\bibitem{ref27}	S. Lan, R. Yu, G. Yu, and L. S. Davis, ``Modeling local geometric structure of 3D point clouds using GeoCNN.'' in \emph{Proceedings of the IEEE Conference on Computer Vision and Pattern Recognition}, 2019, pages 998–1008.
\bibitem{ref28}	Y. Rao, J. Lu, and J. Zhou. ``Spherical fractal convolutional neural networks for point cloud recognition.'' in \emph{Proceedings of the IEEE Conference on Computer Vision and Pattern Recognition}, 2019, pages 452–460.
\bibitem{ref29}	Y. Wang, Y. Sun, Z. Liu, S. E. Sarma, M. M. Bronstein, and J. M. Solomon, ``Dynamic graph cnn for learning on point clouds,'' \emph{ACM
Transactions on Graphics (TOG)}, vol. 38, no. 5, p. 146, 2019.
\bibitem{ref30}	G. Riegler, A. O. Ulusoy, and A. Geiger. ``OctNet: Learning deep 3D representations at high resolutions.'' in \emph{Proceedings of the IEEE Conference on Computer Vision and Pattern Recognition}, 2017, pages 3577–3586.
\bibitem{ref31}	S. Qiu, S. Anwar, and N. Barnes, ``Geometric back-projection network for point cloud classification.'' \emph{IEEE Transactions on Multimedia}, 2021.
\bibitem{ref32}	Z. Wu, Y. Xiong, S. X. Yu, and D. Lin. ``Unsupervised feature learning via non-parametric instance discrimination.'', In \emph{CVPR}, 2018.
\bibitem{ref33}	X. Chen, H. Fan, R. Girshick, and K. He. ``Improved baselines with momentum contrastive learning.'' \emph{arXiv preprint arXiv}:2003.04297, 2020.
\bibitem{ref34}	G. Larsson, M. Maire, and G. Shakhnarovich. ``Learning representations for automatic colorization.'' In \emph{ECCV}, 2016.
\bibitem{ref35}	S. Gidaris, P. Singh, and N. Komodakis, ``Unsupervised representation learning by predicting image rotations.'' In \emph{ICLR}, 2018.
\bibitem{ref36}	C. Doersch, A. Gupta, and A. A Efros. ``Unsupervised visual representation learning by context prediction.'' In \emph{ICCV}, 2015.
\bibitem{ref37}	M. Noroozi and P. Favaro. ``Unsupervised learning of visual representations by solving jigsaw puzzles.'' In \emph{ECCV}, 2016.
\bibitem{ref38}	J. Donahue, and K. Simonyan, ``Large scale adversarial representation learning.'' \emph{arXiv preprint arXiv}:1907.02544, 2019.
\bibitem{ref39}	X. Chen, Y. Duan, R. Houthooft, J. Schulman, I. Sutskever, and P. Abbeel, ``Infogan: Interpretable representation learning by information maximizing generative adversarial nets.'' In \emph{Proceedings of the 30th International Conference on Neural Information Processing Systems}, 2016, pp: 2180-2188.
\bibitem{ref40}	I. Goodfellow, J. Pouget-Abadie , M. Mirza, B. Xu, D. Warde-Farley, S. Ozair, A. Courvile, and Y. Bengio, ``Generative adversarial nets.'' \emph{Advances in neural information processing systems}, 2014, 24.
\bibitem{ref41}	Z. Jure, J. Li, M. Ishan, L. Yann and D. Stéphane, ``Barlow Twins: Self-Supervised Learning via Redundancy Reduction.'', \emph{arXiv preprint arXiv}:2103.03230, 2021.
\bibitem{ref42}	T. Gao,  X. Yao, and D. Chen, ``Simcse: simple contrastive learning of sentence embeddings.'' \emph{arXiv preprint arXiv}:2104.08821, 2021.
\bibitem{ref43}	Y. Zhao, G. Wang, C. Luo, W. Zeng, and Z. J. Zha, ``Self-Supervised Visual Representations Learning by Contrastive Mask Prediction'', in \emph{ICCV}, 2021.
\bibitem{ref44}	N. Zhao, Z. Wu, R.Lau, S. Lin, ``What Makes Instance Discrimination Good For Transfer Learning?'', \emph{arXiv preprint arXiv}: 2006.06606, 2021.
\bibitem{ref45}	I. Misra and L. Maaten. ``Self-supervised learning of pretext-invariant representations'', \emph{arXiv preprint arXiv}: 1912.01991, 2019.
\bibitem{ref46}	R. Qian, T. Meng, B. Gong, M. H. Yang, H. Wang, S. Belongie, Y. Cui, ``Spatiotemporal Contrastive Video Representation Learning'', \emph{arXiv preprint arXiv}: 2008.03800, 2021.
\bibitem{ref47}	G. Lorre, J. Rabarisoa, A. Orcesi, S. Ainouz, and S. Canu. ``Temporal contrastive pretraining for video action recognition.'' In \emph{The IEEE Winter Conference on Applications of Computer Vision}, pp: 662–670, 2020.
\bibitem{ref48}	M. Noroozi and P. Favaro. ``Unsupervised learning of visual representations by solving jigsaw puzzles.'' In \emph{European Conference on Computer Vision}, Springer, pp. 69–84. 2016.
\bibitem{ref49}	M. Cimpoi, S. Maji, I. Kokkinos, S. Mohamed, and A. Vedaldi. ``Describing textures in the wild.'' In \emph{Proceedings of the IEEE Conference on Computer Vision and Pattern Recognition}, pp. 3606–3613, 2014.
\bibitem{ref50}	T. Wang, P. Isola, ``Understanding Contrastive Representation Learning through Alignment and Uniformity on the Hypersphere'', \emph{arXiv preprint arXiv}: 2005.10242, 2020.
\bibitem{ref51}	R. Tang, C. Ma , W. E. Zhang, Q. Wu , and X. Yang, ``Semantic Equivalent Adversarial Data Augmentation for Visual Question Answering'', \emph{arXiv preprint arXiv}: 2007.09592, 2020.
\bibitem{ref52}	F. Schroff, D. Kalenichenko, and J. Philbin, ``FaceNet: A Unified Embedding for Face Recognition and Clustering.'' \emph{IEEE Conference on Computer Vision \& Pattern Recognition}, 2015, vol. 7, no. pp. 815-823.
\bibitem{ref53}	H. O. Song, Y. Xiang , S. Jegelka; and S. Savarese, ``Deep metric learning via lifted structured feature embedding.'' In \emph{CVPR}, 2016.
\bibitem{ref54}	H. Raia, C. Sumit , and L. Yann. ``Dimensionality reduction by learning an invariant mapping.'' In \emph{CVPR}, 2006.
\bibitem{ref55}	X. Wang, and A. Gupta, ``Unsupervised learning of visual representations using videos.'' In \emph{ICCV}, 2015.
\bibitem{ref56}	E. Simo-Serra, E. Trulls, L. Ferraz, I. Kokkinos, P. Fua, and F. Moreno-Noguer, ``Discriminative learning of deep convolutional feature point descriptors.'' In \emph{ICCV}, 2015.
\bibitem{ref57}	C Huang, C. C Loy, and X. Tang, ``Local similarityaware deep feature embedding.'' In \emph{NIPS}, 2016.
\bibitem{ref58}	Y. Yuan, K. Yang, and C. Zhang, ``Hard-aware deeply cascaded embedding.'' In \emph{ICCV}, 2017.
\bibitem{ref59}	H. Shi, Y. Yang, X. Zhu, S. Liao, Z. Lei, W. Zheng, and S. Z. Li, ``Embedding deep metric for person reidentification: A study against large variations.'' In \emph{ECCV}, 2016.
\bibitem{ref60}	Y. Cui, F. Zhou, Y. Lin; and S. Belongie, ``Finegrained categorization and dataset bootstrapping using deep metric learning with humans in the loop.''  In \emph{CVPR}, 2016.
\bibitem{ref61}	M.A. Uy, Q.H. Pham, B.S. Hua, D.T. Nguyen, S.K. Yeung. ``Revisiting Point Cloud Classification: A New Benchmark Dataset and Classification Model on Real-World Data.'', \emph{arXiv preprint arXiv}: 1908.04616, 2019.
\bibitem{ref62}	I. Loshchilov and F. Hutter, ``Sgdr: Stochastic gradient descent with warm restarts,'' \emph{arXiv preprint arXiv}:1608.03983, 2016.
\bibitem{ref63}	Y. Ben-Shabat, M. Lindenbaum, and A. Fischer, ``3dmfv: Threedimensional point cloud classification in real-time using convolutional neural networks,'' \emph{IEEE Robotics and Automation Letters}, vol. 3, no. 4, pp. 3145–3152, 2018.
\bibitem{ref64}	O. J. Henaff, A. Razavi, C. Doersch, S. Eslami, and A. Oord. ``Data-efficient image recognition with contrastive predictive coding.'' \emph{arXiv preprint arXiv}:1905.09272, 2019.
\bibitem{ref65}	Y. Tian, D. Krishnan, and P. Isola. ``Contrastive multiview coding.'' \emph{arXiv preprint arXiv}:1906.05849, 2019.
\bibitem{ref66}	Kaiming He, Haoqi Fan, Yuxin Wu, Saining Xie, and Ross Girshick. ``Momentum contrast for unsupervised visual representation learning.'' \emph{arXiv preprint arXiv}:1911.05722, 2019.
\bibitem{ref67}	D. E. Rumelhart, G.E. Hinton, and R. J. Williams, ``Learning representations by back-propagating errors.'' In \emph{nature}, vol.323, no. 6088, pp. 533–536, 1986.
\bibitem{ref68}	E. B. Baum and F. Wilczek, ``Supervised learning of probability distributions by neural networks.'' In \emph{Neural information processing systems}, pp 52–61, 1988.
\bibitem{ref69}	E. Levin and M. Fleisher. ``Accelerated learning in layered neural networks.'' In \emph{Complex systems}, vol. 2, pp. 625–640, 1988.
\bibitem{ref70}	O. Chapelle, and A. Zien, ``Semi-supervised learning by low density separation.'' In \emph{AISTATS}, pp. 57–64, 2005.
\bibitem{ref71}	Y. Wen, K. Zhang, Z. Li, Y. Qiao, ``A Discriminative Feature Learning Approach for Deep Face Recognition.'' In \emph{ECCV}, vol. 9911, pp. 499-515.2016.
\bibitem{ref72}	L. van der Maaten, G.Hinton, ``Visualizing data using t-SNE'', in \emph{Journal of Machine Learning Reaserch}, vol. 9, no. 2008 , pp. 2579-2605, 2008.
\bibitem{ref73}	E. Shannon, Claude, ``A Mathematical Theory of Communication''. In \emph{Bell System Technical Journal}. Vol. 27, no.3, pp. 379–423, 1948.
\bibitem{ref74}	K. Sohn, ``Improved deep metric learning with multi-class n-pair loss objective.'' In \emph{Advances in neural information processing systems}. pp. 1857-1865,2016.
\bibitem{ref75}	J. Wang, Y.Song, T. Leung, C. Rosenberg, J.  Philbin, B. Chen, and Y. Wu, ``Learning finegrained image similarity with deep ranking.'' In \emph{CVPR}, 2014.
\bibitem{ref76}	X.Wang, Y. Hua, E. Kodirov, G. Hu, and N.M. Robertson, ``Deep metric learning by online soft mining and class-aware attention.'' In \emph{Proceedings of the AAAI Conference on Artificial Intelligence}. vol. 33, No. 01, pp. 5361-5368, 2019.
\bibitem{ref77}	X. Yan, R. Chen, L. Feng, J. Yang, H. Zheng, and W. Zhang, ``Progressive Representative Labeling for Deep Semi-Supervised Learning.'' \emph{arXiv preprint arXiv}:2108.06070, 2021.
\bibitem{ref78}	W. Wu, Z. Qi, L. Fuxin, ``Pointconv: Deep convolutional networks on 3d point clouds.'' \emph{Proceedings of the IEEE/CVF Conference on Computer Vision and Pattern Recognition}, pp. 9621-9630, 2019.
\bibitem{ref79}	X. Wang, X. Han, W. Huang, D. Dong, and M.R. Scott, ``Multi-similarity loss with general pair wrighting for deep metric learning. '' \emph{Proceedings of the IEEE/CVF Conference on Computer Vision and Pattern Recognition}, pp. 5022-5030, 2019.

\end{thebibliography}
\end{document}

\markboth{F.~Yang \lowercase{et al}.: Contrastive Embedding Distribution Refinement and Entropy-Aware Attention for 3D Point Cloud Classification}{F.~Yang et al.: Contrastive Embedding Distribution Refinement and Entropy-Aware Attention for 3D Point Cloud Classification}

\title{SUPPLEMENTARY MATERIAL}
\maketitle

\subsection{Implementation Details}
Our method is a novel complement to the conventional classification method based on a softmax classifier. After the backbone extracts the global feature of the samples, we feed it to the projection head to map to embedding space. We adapt a feedforward neural network with the same number of neurons as a global feature in the projection head. For example, after the GBNet backbone extracts a 2048-dimensional representation feature, we add a MLP (2048) layer that converts it to embedding space. Our contrastive loss is calculated based on samples in a mini-batch. The CPCM computes the distance of embedding feature clusters to assign weights to the samples of different classes to make confusion-prone category pairs yield greater influence on contrast loss. The EAA assigned sample pairs weights based on the weights of special samples. We train our models for 300 epochs with the SGD optimizer with a momentum of 0.9. The learning rate is decreased from 0.1 to 0.001 by cosine annealing.

\subsection{Integration of the CPCM and EAA}
Our approach is implementing the embedding distribution refinement that combines the CPCM and EAA that both can be regarded as the samples reweight mechanism in mini-batch. The weights of the former module are class-based calculations, while the latter are sample-based. Thus the combination of the two modules in our approach is direct recombination of the weights. Assume we obtain the weight of class pairs in CPCM defines in eq.4:
\begin{equation}
W_{C P C M}^{-}(i, j)=\frac{e^{\operatorname{dist}(C(i), C(j))}+e^{-\operatorname{dist}(C(i), C(j))}}{e^{\operatorname{dist}(C(i), C(j))}}
\end{equation}
And the EAA module on the definition of the weights of sample pairs as eq.8:
\begin{equation}
W_{E A A}(i, j)=\operatorname{select}\left(a_{i}, a_{j}\right).
\end{equation}

We hypothesized that these two weights are jointly applied to the sample with orthogonal directions. Accordingly we employ the parallelogram rule of weight space to synthesize these two forces. The Integration of the two types of weights in our approach can be deduced to:
\begin{equation}
W_{\text {overall }}(i, j)=\sqrt{W_{C P C M}^{-}(i, j)^{2}+W_{E A A}(i . j)^{2}}
\end{equation}

\begin{table}[htbp]
  \centering
  \caption{Various Loss With Our Supervised Contrastive Classification On the ScanObjectNN With A GBNet Backbone.}
    \begin{tabular}{ccc}
    \toprule
    \toprule
    \multicolumn{1}{c}{Supervised contrastive classification} & Avg. class Acc. & Overall Acc. \\
    \midrule
      InfoNCEloss [3]   & 78.4  & 81.4 \\
      Multi-Similarity loss [79] & 74.3  & 78.8 \\
    \bottomrule
    \bottomrule
    \end{tabular}%
  \label{tab:addlabel}%
\end{table}%

\begin{figure}
\centering
\includegraphics[width=8cm]{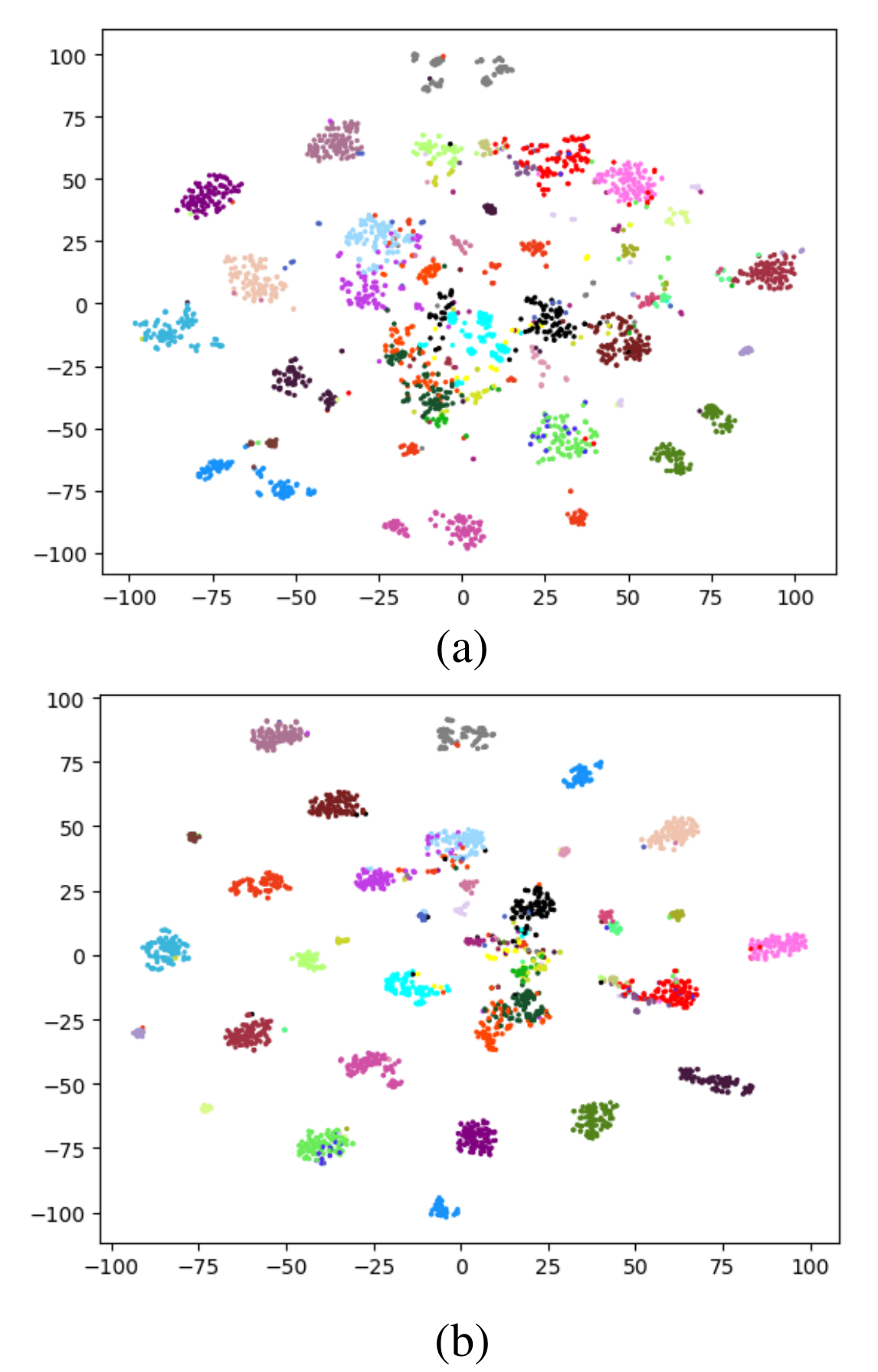}
\caption{T-SNE visualization of point cloud classification for the PointNet++ backcbone on the ModelNet40 dataset. (a) baseline, (b) our approach.}
\end{figure}

\subsection{Impact of the constrative loss in SCC}
In addition to the popularly InfoNCE loss was used to implement the . We also applied other metric losses in deep metric learning for comparison experiments. The results are as follows.

We investigate the influence of these two modeling embeddings similarity loss on point cloud classification. As shown in TABLE \uppercase\expandafter{\romannumeral1}, the InfoNCEloss worked better, probably because each anchor embedding was compared and refinemented with more positive and negative sample.

\subsection{Feature Visualization on ModelNet40}
We visualize our learned features by t-SNE on Modelnet40 in Fig. 1. As shown in Fig. 1, the learned feature by our approach is more compact and clearly bounded than the baseline with only softmax loss training.